\documentclass{article}

    \PassOptionsToPackage{numbers, compress}{natbib}
\usepackage[main, final]{neurips_2026}

\usepackage{wrapfig}
\usepackage{arydshln}
\usepackage[utf8]{inputenc} 
\usepackage[T1]{fontenc}    
\usepackage{hyperref}       
\usepackage{url}            
\usepackage{booktabs}       
\usepackage{amsfonts}       
\usepackage{nicefrac}       
\usepackage{microtype}      
\usepackage[table,dvipsnames]{xcolor} 
\usepackage{enumitem}              
\usepackage{graphicx}
\usepackage{amsmath, amssymb, bm}
\usepackage{algorithm}      
\usepackage{algpseudocode} 
\usepackage{wrapfig}
\usepackage{array}

\usepackage{multirow}
\definecolor{lightbluepurple}{RGB}{231,230,245}

\title{RaPD: Resolution-Agnostic Pixel Diffusion via Semantics-Enriched Implicit Representations}

\author{%
  Yanhao Ge\textsuperscript{1} \quad Shanyan Guan\textsuperscript{2} \quad Weihao Wang\textsuperscript{1,2} \quad Ying Tai\textsuperscript{3} \quad Mingyu You\textsuperscript{1,4\thanks{\noindent M. You is the corresponding author with email: myyou@tongji.edu.cn.}} \\
\textsuperscript{1} College of Electronic and Information Engineering, Tongji University\\
\textsuperscript{2} vivo Mobile Communication Co., Ltd. ~
\textsuperscript{3} Nanjing University\\
\textsuperscript{4}State Key Laboratory of Autonomous Intelligent Unmanned Systems, \\
Frontiers Science Center for Intelligent Autonomous Systems, Ministry of Education\\
}

\begin{document}

\maketitle

\begin{abstract}
Natural images are continuous, yet most generative models synthesize them on discrete grids, limiting resolution-flexible generation. Continuous neural fields enable resolution-free rendering, but prior methods introduce continuity only at the decoding stage as an interpolation module, leaving the generative latent space discretized and reconstruction-oriented. We propose \textbf{RaPD} (\textbf{R}esolution-\textbf{a}gnostic \textbf{P}ixel \textbf{D}iffusion), which performs diffusion in a continuous Neural Image Field (NIF) latent space. RaPD bridges this \textit{reconstruction--generation gap} with \emph{Semantic Representation Guidance} for generation-aware latent learning and a \emph{Coordinate-Queried Attention Renderer} for coordinate-conditioned, scale-aware rendering. A single denoised latent can be rendered at arbitrary resolutions by changing only the query coordinates, keeping diffusion cost fixed. Experiments demonstrate superior generation quality and resolution scalability.

\end{abstract}

\section{Introduction}
\label{sec:intro}

\begin{wrapfigure}{r}{0.45\textwidth}
    \centering
    \vspace{-1.25em}
    \includegraphics[width=\linewidth]{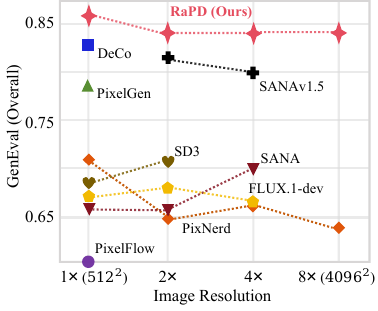}
    \vspace{-0.6cm}
    \caption{
    %
    RaPD supports arbitrary-resolution text-to-image generation and outperforms strong latent- and pixel-diffusion baselines ~\cite{Esser_SD3_2024,flux2024,ma2026pixelgen, Ma_DeCo_2025, Wang_PixNerd_2025, Chen_PixelFlow_2025}.
    }
    \label{fig:teaser}
    \vspace{-0.75em}
\end{wrapfigure}

The visual world is inherently continuous, yet modern image generative models mostly operate on spatially discretized representations. Whether in VAE latent or pixel space~\cite{Rombach_LDM_2022, Esser_SD3_2024, flux2024, Ho_DDPM_2020, Peebles_Xie_2022}, their architectures and computation are tied to discrete grids, making higher resolutions require more tokens and compute~\cite{Zhao_UltraImage_2025}. As a result, native generation at arbitrary, unseen resolutions remains challenging. In this work, we ask whether image generation can instead be formulated over ``continuous pixels'', rather than discrete pixels or compressed VAE proxies.

Existing arbitrary-resolution generation models typically introduce continuity only at the rendering stage. GAN models use coordinate-queried MLPs but suffer from limited generation capacity~\cite{Anokhin_CIPS_2021, INRGAN2021, Chai_ArbitraryScale_2022}. Latent diffusion-based models use flexible decoders~\cite{lu2024fit, Wang_FiTv2_2024, Gao_ArbitraryScale_2023, Chen_INFD_2024, InfGen_2025, Wang_Bai_Yue_Ouyang_Zhang_2025}, yet diffusion itself remains in a discrete space, which ties semantics to a discrete contents and inheriting the VAE bottleneck~\cite{Kingma_VAE_2014, Chen_DCAE_2025, Yao_VAVAE_2025, Leng_REPA-E_2025, yu2025REPA, Wu_Zhang_Shi_Gao_Chen_Wang_Chen_Gao_Tang_Yang_2025}. Pixel diffusion removes the VAE~\cite{Chen_PixelFlow_2025, Yu_PixelDiT_2025, Chen_DiP_2025, Ma_DeCo_2025, Li_He_JiT_2026, ma2026pixelgen, Crowson_HDiT_2024, EPG_NoVAE_2026, chen2026l2p}, but still generates on discrete grids. Even PixNerd~\cite{Wang_PixNerd_2025}, the closest attempt, uses a patch-wise neural field mainly as an interpolation module over fixed-resolution features, rather than learning a latent that models continuous image signals for generation. This motivates a more fundamental question: can we directly generate continuous image signals?

A straightforward solution is to diffuse existing continuous image representations. As shown in Fig.~\ref{fig:nif_compare}, however, this yields poor results. The bottleneck lies in the representation itself: methods such as LIIF~\cite{chen_liif_2020} and CLIF~\cite{Chen_INFD_2024} are built for interpolation, with reconstruction training that favors local appearance cues and lightweight decoders with limited semantic aggregation. Thus, continuous representations must be made generation-aware to support high-quality arbitrary-resolution synthesis.

To address this gap, we propose \textbf{RaPD}, a framework for directly generating continuous image signals. RaPD performs pixel-space diffusion in a continuous Neural Image Field (NIF) latent space, rather than treating neural fields merely as decoding-time interpolation modules. Since existing continuous image representations are reconstruction-oriented, RaPD first reshapes the NIF latent into a generation-aware representation through \emph{Semantic Representation Guidance} (SRG), which injects semantic structure from a vision foundation model and adapts the timestep schedule to high-density continuous latents. To decode such semantically enriched latents, RaPD further introduces a \emph{Coordinate-Queried Attention Renderer} (CQAR), which replaces local implicit decoders with coordinate-conditioned, scale-aware attention for inter-pixel semantic reasoning. At inference, a single denoised NIF latent can be rendered at arbitrary target resolutions by varying only the query coordinate grid, while the diffusion cost remains independent of output size (shown in Fig.~\ref{fig:teaser}).

Our contributions are summarized as follows:
\vspace{-0.1cm}
\begin{enumerate}[ leftmargin=1.5em, nosep]
    \item We propose RaPD, a resolution-agnostic pixel diffusion framework that directly generates continuous image signals in an NIF latent space.
    \item We introduce {Semantic Representation Guidance} to transform reconstruction-oriented continuous latents into generation-aware representations.
    \item We design a {Coordinate-Queried Attention Renderer} for spatially coherent arbitrary-resolution rendering with fixed diffusion cost.
\end{enumerate}

\vspace{-0.9em}
\section{Preliminaries}
\label{sec:prelim}

\paragraph{Flow Matching.}
Flow matching~\cite{lipman2023flowmatching, Esser_SD3_2024} is a widely used training paradigm for diffusion models~\cite{sohldickstein2015DiffusionModel, Ho_DDPM_2020, song2021ScoreMatching}. It learns a velocity field that transports Gaussian noise to the data distribution. To avoid overloading image notation, we denote a generic flow state by \(\boldsymbol{u}_t\). For a data sample \(\boldsymbol{u}_1\), a common linear interpolation path is
\begin{equation}
  \boldsymbol{u}_t
  =
  (1-t)\boldsymbol{\epsilon}
  +
  t\boldsymbol{u}_1,
  \quad
  \boldsymbol{\epsilon}\sim\mathcal{N}(\boldsymbol{0},\boldsymbol{I}),
  \label{eq:fm_forward}
\end{equation}
where \(t\in[0,1]\). Under this convention, \(\boldsymbol{u}_t\) evolves from Gaussian noise at \(t=0\) to data at \(t=1\), and the target velocity is
\(\boldsymbol{v}^{\star}=\boldsymbol{u}_1-\boldsymbol{\epsilon}\).
For image generation, \(\boldsymbol{u}_1\) is typically instantiated as a spatially discretized representation, such as VAE latents~\cite{Rombach_LDM_2022, Podell_SDXL_2023} or image patch tokens~\cite{Chen_PixelFlow_2025, Yu_PixelDiT_2025, Chen_DiP_2025, Li_He_JiT_2026, ma2026pixelgen, Crowson_HDiT_2024, EPG_NoVAE_2026}. Despite their strong performance, these representations tie the learned velocity field to finite spatial grids and token lengths, which can limit generalization to unseen resolutions~\cite{bu2025hiflowtrainingfreehighresolutionimage, he2023scalecraftertuningfreehigherresolutionvisual, du2023demofusiondemocratisinghighresolutionimage}.

\vspace{-1em}
\paragraph{Neural Image Fields.}
We use \(\boldsymbol{x}\) to denote an RGB image and \(\boldsymbol{z}\) to denote its latent feature representation. Beyond discrete image grids, an image can be represented as a continuous coordinate-based function, commonly referred to as a Neural Image Field (NIF)~\cite{Sitzmann_INR_2020, mildenhall2020NERF}. Representative methods such as LIIF~\cite{chen_liif_2020} and CLIF~\cite{Chen_INFD_2024} parameterize an image by a feature map \(\boldsymbol{z}\in\mathbb{R}^{C\times H\times W}\), and render RGB values by querying continuous coordinates:
\begin{equation}
    \hat{\boldsymbol{x}}(\boldsymbol{q})
    =
    f_{\psi}
    \big(
    \boldsymbol{z};\boldsymbol{q},\Delta\boldsymbol{c}
    \big),
    \label{eq:nif_render}
\end{equation}
where \(\boldsymbol{q}\in[-1,1]^2\) is a target coordinate, \(\Delta\boldsymbol{c}\) is the cell-size descriptor associated with the target pixel, and \(f_{\psi}\) denotes a coordinate renderer that queries features from \(\boldsymbol{z}\) and predicts the RGB value at \(\boldsymbol{q}\). By evaluating Eq.~\eqref{eq:nif_render} over different coordinate grids, the same representation can be rendered at arbitrary resolutions.
Existing NIFs are typically trained for interpolation or reconstruction. Given a target RGB image \(\boldsymbol{x}_{\mathrm{tar}}\) defined on a coordinate grid \(\mathcal{Q}_{\mathrm{tar}}\), the renderer produces
\(\hat{\boldsymbol{x}}_{\mathrm{tar}}
=
\{f_{\psi}(\boldsymbol{z};\boldsymbol{q},\Delta\boldsymbol{c}(\boldsymbol{q}))\mid \boldsymbol{q}\in\mathcal{Q}_{\mathrm{tar}}\}\).
The reconstruction objective is
\begin{equation}
    \label{eq:nif-trainloss}
    \mathcal{L}_{\mathrm{rec}}
    =
    \left\|
    \hat{\boldsymbol{x}}_{\mathrm{tar}}
    -
    \boldsymbol{x}_{\mathrm{tar}}
    \right\|_1
    +
    \omega
    \mathcal{L}_{\mathrm{LPIPS}}
    \big(
    \hat{\boldsymbol{x}}_{\mathrm{tar}},
    \boldsymbol{x}_{\mathrm{tar}}
    \big),
\end{equation}
where \(\omega\) is the LPIPS loss weight. This reconstruction-oriented objective makes NIFs effective interpolators, but their latents are not directly generation-ready.

\section{Resolution-Agnostic Pixel Diffusion}
\label{sec:method}

\begin{figure}[t]
\centering
\includegraphics[width=\columnwidth]{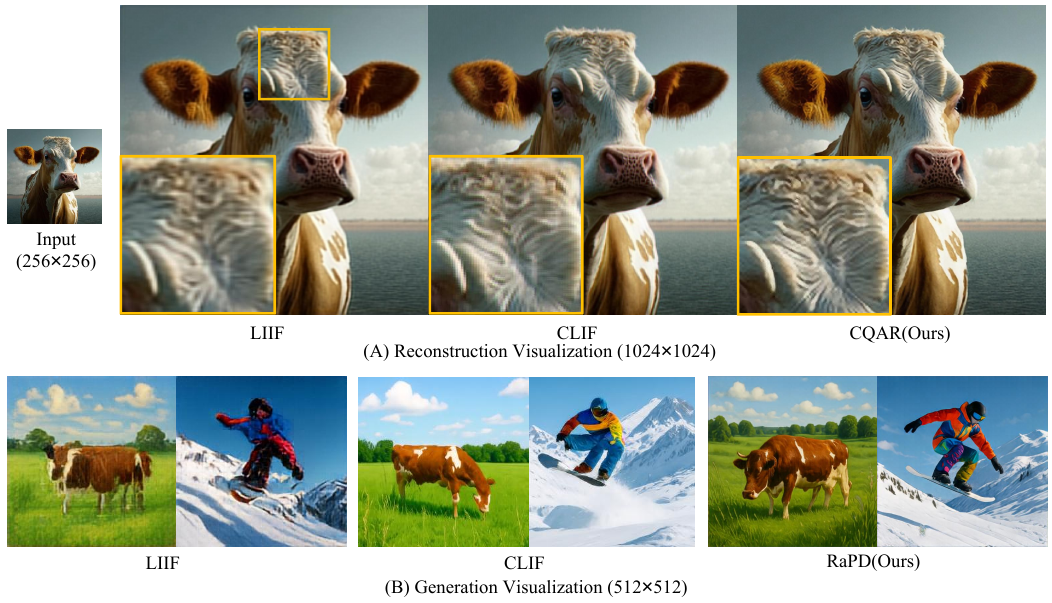}
\vspace{-0.3cm}
\caption{Reconstruction--generation gap of existing NIF representations: strong super-resolution reconstruction (top) but weak text-to-image generation (bottom).}
\label{fig:nif_compare}
\end{figure}

We present \textbf{RaPD}, a framework for text-to-image generation over continuous NIF latents. Instead of generating a discrete image grid or a fixed-resolution VAE latent, RaPD learns a generation-aware continuous NIF latent and denoises it with flow matching. The denoised latent is then rendered at arbitrary target resolutions by querying a coordinate-conditioned renderer, keeping the diffusion cost independent of output size.

A straightforward baseline is to apply a PixelGen-style DiT denoiser~\cite{Peebles_Xie_2022, ma2026pixelgen} to existing NIF latents, such as LIIF~\cite{chen_liif_2020} and CLIF~\cite{Chen_INFD_2024}. As shown in Fig.~\ref{fig:nif_compare}, these latents reconstruct well but generate poorly, exposing a \emph{reconstruction--generation gap}. RaPD addresses this gap with two components: \emph{Semantic Representation Guidance} (SRG), which makes NIF latents generation-aware, and \emph{Coordinate-Queried Attention Renderer} (CQAR), which renders queried coordinates with inter-pixel semantic reasoning.

\vspace{-0.8em}
\subsection{Basic Framework}
\label{sec:framework}
\vspace{-0.5em}

\begin{figure*}[t]
  \centering
  \includegraphics[width=0.95\linewidth]{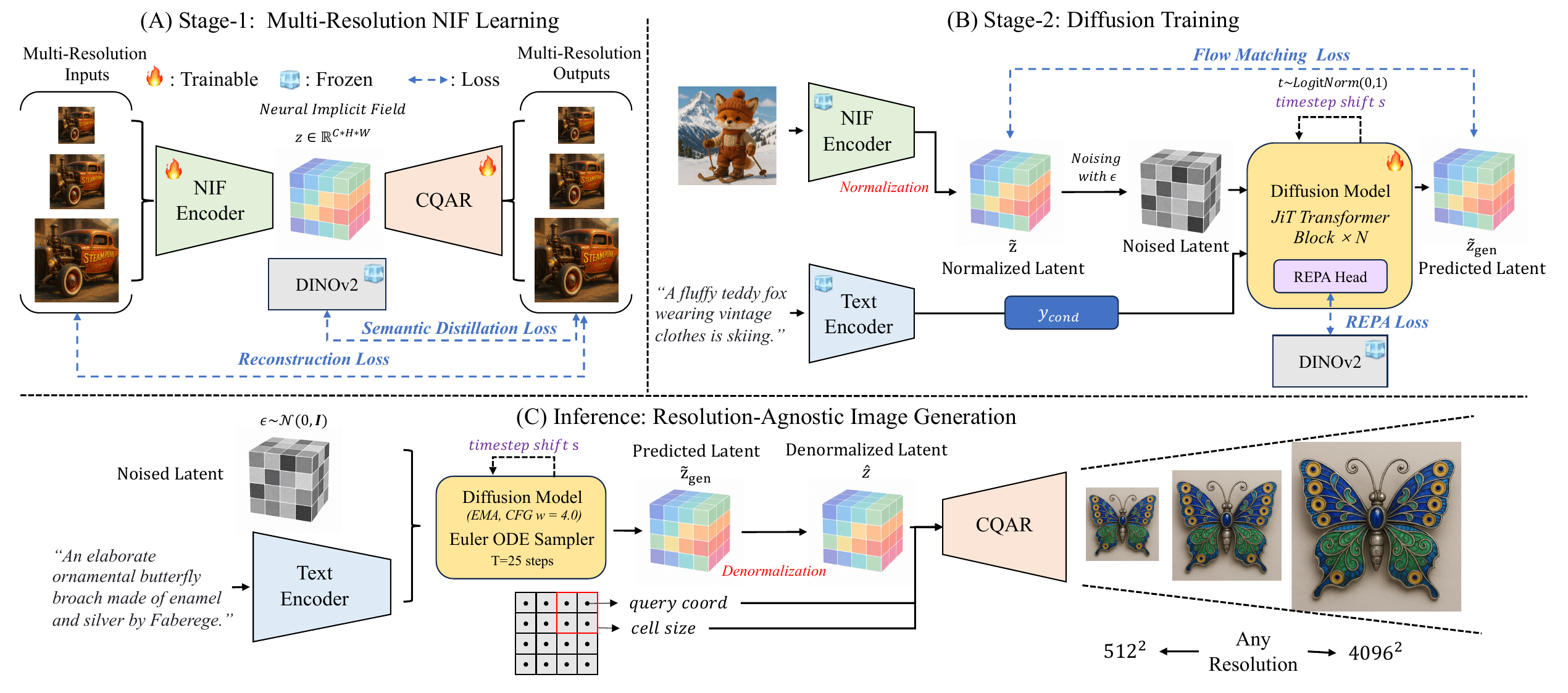}
  \caption{Overview of RaPD. (A) Stage-1: semantically guided multi-resolution NIF learning. (B) Stage-2: text-conditioned flow matching in normalized NIF latents. (C) Inference: render one denoised latent at arbitrary resolutions via CQAR.}
  \vspace{-0.4cm}
  \label{fig:framework}
\end{figure*}

As shown in Fig.~\ref{fig:framework}, RaPD adopts a two-stage design. Stage-1 learns a generation-aware NIF latent. Given an encoder input image \(\boldsymbol{x}_{\mathrm{in}}\), an encoder \(E_{\phi}\) maps it to a dense latent
\(\boldsymbol{z}=E_{\phi}(\boldsymbol{x}_{\mathrm{in}})\in\mathbb{R}^{C\times H\times W}\)
without spatial downsampling. Given a target coordinate \(\boldsymbol{q}\), CQAR renders the corresponding RGB value as
\begin{equation}
    \hat{\boldsymbol{x}}(\boldsymbol{q})
    =
    f_{\psi}^{\mathrm{CQAR}}
    \big(
    \boldsymbol{z};\boldsymbol{q},\Delta\boldsymbol{c}
    \big),
    \label{eq:rapd_render}
\end{equation}
where \(f_{\psi}^{\mathrm{CQAR}}\) queries local features from \(\boldsymbol{z}\) at coordinate \(\boldsymbol{q}\), and \(\Delta\boldsymbol{c}\) is the cell-size used by CQAR. An image at resolution \(H'\times W'\) is rendered by evaluating Eq.~\eqref{eq:rapd_render} over all target coordinates. Unlike conventional NIFs trained mainly for interpolation, RaPD trains \(E_{\phi}\) and \(f_{\psi}^{\mathrm{CQAR}}\) with SRG so that \(\boldsymbol{z}\) preserves coordinate-rendering capability while acquiring semantic structure useful for generation.

We instantiate \(E_{\phi}\) with an EDSR-style encoder~\cite{lim2017EDSR,chen_liif_2020} without spatial downsampling, producing a dense latent at the encoder-input resolution. 
In Stage-2, \(E_{\phi}\) and \(f_{\psi}^{\mathrm{CQAR}}\) are frozen. We compute per-channel statistics \(\boldsymbol{\mu}_z\) and \(\boldsymbol{\sigma}_z\) over the training set and train a DiT-style denoiser \(D_{\theta}\), following the PixGen design~\cite{ma2026pixelgen}, over normalized NIF latents
\(\tilde{\boldsymbol{z}}=(\boldsymbol{z}-\boldsymbol{\mu}_z)/\boldsymbol{\sigma}_z\).
The denoiser is conditioned on frozen Qwen3-1.7B~\cite{Yang_Li_Yang_Zhang_Hui_Zheng_Yu_Gao_Huang_Lv_2025} text embeddings
\(\boldsymbol{y}_{\mathrm{cond}}\in\mathbb{R}^{128\times2048}\)
and trained to predict the flow-matching velocity. At inference, integrating the learned velocity field produces a generated normalized latent \(\tilde{\boldsymbol{z}}_{\mathrm{gen}}\), which is denormalized and rendered by CQAR at any requested resolution.

\vspace{-0.8em}
\subsection{Semantic Representation Guidance}
\label{sec:srg}
\vspace{-0.5em}

Existing NIF latents are optimized for interpolation and reconstruction, making them poorly suited for generative modeling. We propose \emph{Semantic Representation Guidance} (SRG), a unified strategy that makes NIF latents generation-aware at both the representation and diffusion-training levels. SRG consists of two components: \emph{Semantic Distillation}, which injects foundation-model semantics into the dense NIF latent during Stage-1, and \emph{Latent-Density-Aware Timestep Shift}, which adapts Stage-2 flow matching to the high-density SRG-shaped latent.

\vspace{-1em}
\paragraph{Semantic Distillation.}
How can a NIF latent be made generation-aware while retaining its ability to render continuous coordinates? We argue that the latent should not be learned from pixel reconstruction alone, but should also inherit semantic structure from a vision foundation model. We adopt a frozen DINOv2 ViT-B/14~\cite{Oquab_Darcet_Moutakanni_2024} as the semantic teacher, producing patch features
\(\boldsymbol{t}\in\mathbb{R}^{768\times H_p\times W_p}\).
Since the NIF latent \(\boldsymbol{z}\) has no spatial downsampling and a different channel dimension, we pool \(\boldsymbol{z}\) to the DINOv2 patch grid and project it with \(1\times1\) convolution, yielding student features
\(\boldsymbol{z}_{\mathrm{proj}}\in\mathbb{R}^{768\times H_p\times W_p}\).
This projection branch is used only for distillation; the original NIF latent keeps its spatial layout and channel dimension for coordinate-based rendering.

The alignment loss comprises two terms that capture both local semantic content and global relational structure~\cite{yu2025REPA, Yao_VAVAE_2025, Wu_Zhang_Shi_Gao_Chen_Wang_Chen_Gao_Tang_Yang_2025}.
%
%
First, a cosine loss enforces per-position semantic agreement up to a margin:
\begin{equation}
  \mathcal{L}_{\mathrm{mcos}}
  =
  \frac{1}{N}
  \sum_{i=1}^{N}
  \max\bigl(
  0,\,
  1 - m_{\mathrm{cos}}
  - \cos(\boldsymbol{z}_{\mathrm{proj}}^{(i)},\boldsymbol{t}^{(i)})
  \bigr),
  \label{eq:mcos}
\end{equation}
where \(N=H_p\times W_p\). Second, a distance-matrix similarity loss preserves the teacher's pairwise relational geometry:
\begin{equation}
  \mathcal{L}_{\mathrm{mdms}}
  =
  \frac{1}{K^2}
  \sum_{i,j}
  \max\bigl(
  0,\,
  |\boldsymbol{D}_z[i,j]-\boldsymbol{D}_t[i,j]|
  -m_{\mathrm{dist}}
  \bigr),
  \label{eq:mdms}
\end{equation}
where \(\boldsymbol{D}_z\) and \(\boldsymbol{D}_t\) are cosine-similarity matrices computed over \(K\) sampled spatial positions from \(\boldsymbol{z}_{\mathrm{proj}}\) and \(\boldsymbol{t}\), respectively. The final semantic distillation loss is
\(\mathcal{L}_{\mathrm{distill}}=\mathcal{L}_{\mathrm{mcos}}+\mathcal{L}_{\mathrm{mdms}}\).
We use \(m_{\mathrm{cos}}=0.5\), \(m_{\mathrm{dist}}=0.25\), and \(K=256\).

Unlike prior vision-feature alignment applied to downsampled VAE latents or denoiser features~\cite{yu2025REPA, Yao_VAVAE_2025, Wu_Zhang_Shi_Gao_Chen_Wang_Chen_Gao_Tang_Yang_2025}, our alignment supervises a dense NIF latent directly. The projection branch transfers semantic structure without changing the latent layout, while the margin-based losses preserve compatibility with pixel-level reconstruction. Thus, semantic alignment acts here as formative supervision for a generation-aware continuous representation, rather than an auxiliary regularizer on an existing latent.

\vspace{-1em}
\paragraph{Latent-Density-Aware Timestep Shift.}
Semantic Distillation shapes the NIF latent, but Stage-2 must still learn a flow over this unusually dense representation. Unlike compressed VAE latents, the NIF latent has no spatial downsampling and retains a high-resolution coordinate-rendering layout. In flow matching, the timestep shift factor \(s\) controls the sampled timestep distribution:
\begin{equation}
    t_0=\mathrm{sigmoid}(\xi),
    \quad
    \xi\sim\mathcal{N}(0,1),
    \qquad
    t=\frac{t_0}{t_0+(1-t_0)\cdot s}.
    \label{eq:timestep_shift}
\end{equation}
%
A larger $s$ helps the model prioritize heavy noise stages (small $t$). These stages are the most critical for generating decent content.
%
Standard VAE-latent diffusion models typically use small shift factors, with \(s\in[1,3]\), since their latents are spatially compressed. For NIF latents, we follow the resolution-dependent shift principle~\cite{Esser_SD3_2024} and adopt a larger shift to compensate for the increased information density of the latent. As analyzed in Sec.~\ref{sec:analysis_experiments}, both insufficient and excessive shifts degrade generation quality; we empirically set \(s=32\) in RaPD. Together, Semantic Distillation and Latent-Density-Aware Timestep Shift form SRG: the former shapes the latent into a semantic representation, while the latter adapts diffusion training to recover this representation during generation.

\begin{figure*}[t]
  \centering
  \includegraphics[width=0.85\linewidth]{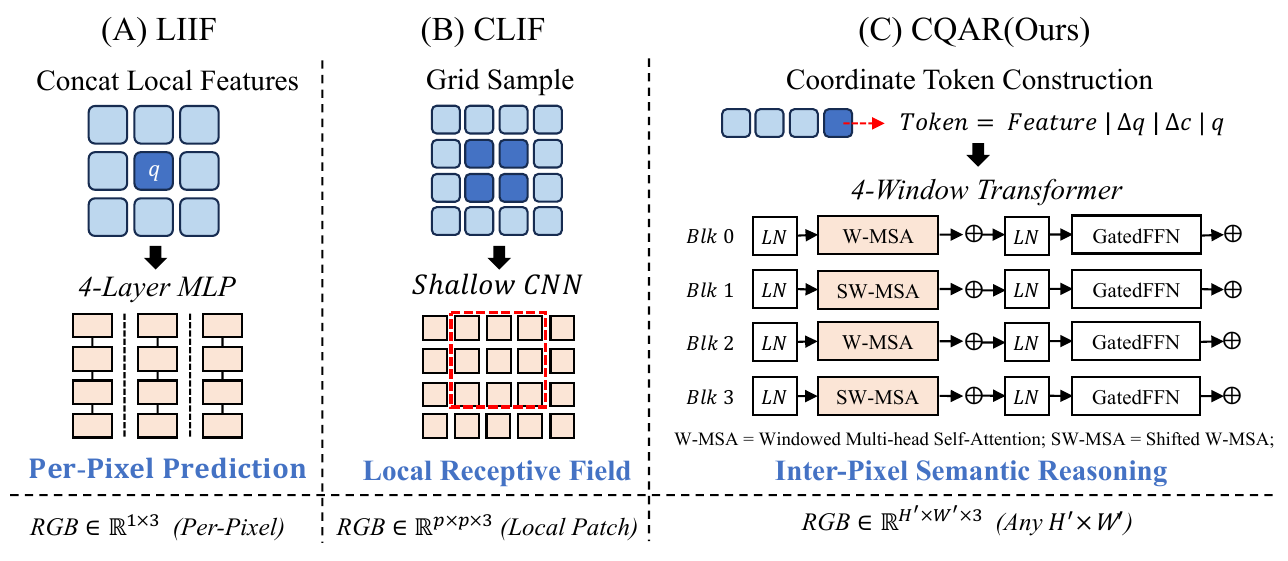}
  \vspace{-0.3cm}
  \caption{Implicit decoder comparison. (A) LIIF: per-pixel MLP. (B) CLIF: shallow CNN. (C) CQAR: coordinate-conditioned windowed attention over output tokens.}
  \vspace{-1em}
  \label{fig:modules}
\end{figure*}

\subsection{Coordinate-Queried Attention Renderer}
\label{sec:cqar}

While SRG provides the NIF latent with semantic structure, the renderer must exploit this structure during coordinate-based decoding. Existing implicit decoders such as LIIF~\cite{chen_liif_2020} and CLIF~\cite{Chen_INFD_2024} predict pixels from local latent features using pointwise MLPs or shallow CNNs. These local operators are effective for interpolation, but cannot reason across queried pixels and therefore underuse the semantic features injected by SRG.

We address this limitation with the \emph{Coordinate-Queried Attention Renderer} (CQAR), shown in Fig.~\ref{fig:modules}. The key distinction of CQAR is that attention is performed over queried output-coordinate tokens, rather than over the input latent grid. For a target resolution \(H'\times W'\), CQAR constructs one token per output coordinate and processes these tokens with Swin-style windowed self-attention~\cite{liu2021swin_transformer}. Each token is built from the queried latent feature, absolute coordinate, relative offset, and target cell size, making the renderer coordinate-grounded and scale-aware. Windowed self-attention over these output tokens enables global inter-pixel reasoning while keeping rendering cost scalable.

\vspace{-1em}
\paragraph{Coordinate Token Construction.}
Given an input latent \(\boldsymbol{z}\in\mathbb{R}^{C\times H\times W}\) and a target coordinate \(\boldsymbol{q}_{i,j}\), CQAR forms an output-coordinate token by
\begin{equation}
  \boldsymbol{h}_{i,j}
  =
  \phi_{\mathrm{in}}
  \big(
  \boldsymbol{z}_{\mathrm{nn}}(\boldsymbol{q}_{i,j}),
  \Delta\boldsymbol{q}_{i,j},
  \Delta\boldsymbol{c}_{i,j},
  \boldsymbol{q}_{i,j}
  \big),
  \label{eq:query}
\end{equation}
where \(\boldsymbol{z}_{\mathrm{nn}}(\boldsymbol{q}_{i,j})\) is the nearest-neighbor latent feature, \(\Delta\boldsymbol{q}_{i,j}=\boldsymbol{q}_{i,j}-\boldsymbol{q}^{*}_{i,j}\) is the offset from the nearest latent-grid center \(\boldsymbol{q}^{*}_{i,j}\), \(\Delta\boldsymbol{c}_{i,j}\) is the relative cell-size descriptor converted from the raw pixel extent, and \(\boldsymbol{q}_{i,j}\in[-1,1]^2\) is the absolute coordinate. The projected tokens are processed by windowed Transformer blocks and mapped to RGB values. Thus, each token attends to neighboring output-coordinate tokens, enabling global inter-pixel reasoning that pointwise implicit decoders lack.

\subsection{Training and Inference}
\label{sec:train_infer}

RaPD is trained in two stages. Stage-1 (Fig.~\ref{fig:framework} (A)) learns a semantically guided NIF latent. Stage-2 (Fig.~\ref{fig:framework} (B)) freezes the NIF encoder and CQAR, then trains a text-conditioned DiT denoiser over normalized NIF latents with flow matching. During inference(Fig.~\ref{fig:framework} (C)), the denoiser generates one normalized NIF latent, and CQAR renders the latent on any target coordinate grid. 

\vspace{-0.8em}
\subsubsection{Training}
\vspace{-0.5em}

\paragraph{Stage-1: Generation-Aware NIF Learning.}
Stage-1 trains the NIF encoder and CQAR to learn a generation-aware continuous latent while preserving coordinate-based renderability. Specifically, the encoder \(E_{\phi}\) maps an input image \(\boldsymbol{x}_{\mathrm{in}}\) to a dense NIF latent \(\boldsymbol{z}=E_{\phi}(\boldsymbol{x}_{\mathrm{in}})\), and CQAR renders this latent over target coordinates. To prevent the latent from being reconstruction-oriented, we optimize it with the following loss:
\begin{equation}
  \mathcal{L}_{\mathrm{Stage1}}
  =
  \mathcal{L}_{\mathrm{rec}}
  +
  w_{\mathrm{adapt}}\mathcal{L}_{\mathrm{distill}},
  \quad
  w_{\mathrm{adapt}}
  =
  w_{\mathrm{base}}
  \cdot
  \mathrm{clamp}
  \left(
  \frac{
  \|\nabla_{\phi_L}\mathcal{L}_{\mathrm{rec}}\|
  }{
  \|\nabla_{\phi_L}\mathcal{L}_{\mathrm{distill}}\|+\delta_1
  },
  0,10^8
  \right),
  \label{eq:stage1_loss}
\end{equation}
where \(\mathcal{L}_{\mathrm{distill}}\) is the semantic distillation loss defined in Sec.~\ref{sec:srg}, \(\phi_L\) denotes the parameters of the encoder's last convolutional layer, \(\delta_1=10^{-4}\), and \(w_{\mathrm{base}}=0.1\). This adaptive weighting balances the reconstruction and semantic objectives, shaping \(\boldsymbol{z}\) into a generation-aware NIF latent without sacrificing reconstruction fidelity.

We further adopt multi-resolution supervision to expose CQAR to varying output cell sizes. Given a training image, we sample \(r\sim\mathcal{U}[1,2]\) and extract a high-resolution crop of size \((\lfloor rH_0\rceil,\lfloor rW_0\rceil)\) as the target image \(\boldsymbol{x}_{\mathrm{tar}}\), where \((H_0,W_0)\) is the fixed encoder input resolution. We downsample \(\boldsymbol{x}_{\mathrm{tar}}\) to obtain \(\boldsymbol{x}_{\mathrm{in}}\) and encode it into the NIF latent space. The queried coordinates and cell sizes are defined with respect to the full high-resolution crop, so that the same latent is trained to render RGB values under different output cell sizes using per-pixel reconstruction loss in Eq.~\eqref{eq:nif-trainloss}. After Stage-1, per-channel statistics \(\boldsymbol{\mu}_z,\boldsymbol{\sigma}_z\) are computed over training-set latents for Stage-2 normalization.

\vspace{-1em}
\paragraph{Stage-2: Diffusion Training.}
Once finished with semantic distillation, Stage-2 trains a text-conditioned DiT denoiser \(D_{\theta}\) over the learned NIF latent. For each image, we encode \(\boldsymbol{x}_{\mathrm{in}}\) into \(\boldsymbol{z}=E_{\phi}(\boldsymbol{x}_{\mathrm{in}})\) and normalize it as
\(\tilde{\boldsymbol{z}}=(\boldsymbol{z}-\boldsymbol{\mu}_z)/\boldsymbol{\sigma}_z\).
Here the data endpoint \(\boldsymbol{u}_1\) in Eq.~\eqref{eq:fm_forward} is instantiated as \(\tilde{\boldsymbol{z}}\). We sample \(t\) following Eq.~\eqref{eq:timestep_shift}, draw \(\boldsymbol{\epsilon}\sim\mathcal{N}(\boldsymbol{0},\boldsymbol{I})\), and train the denoiser with
\begin{equation}
    \boldsymbol{u}_t
    =
    (1-t)\boldsymbol{\epsilon}
    +
    t\tilde{\boldsymbol{z}},
    \quad
    \boldsymbol{v}_{\theta}
    =
    D_{\theta}
    (
    \boldsymbol{u}_t,
    t,
    \boldsymbol{y}_{\mathrm{cond}}
    ),
    \quad
    \mathcal{L}_{\mathrm{FM}}
    =
    \left\|
    \boldsymbol{v}_{\theta}
    -
    (
    \tilde{\boldsymbol{z}}
    -
    \boldsymbol{\epsilon}
    )
    \right\|_2^2.
    \label{eq:stage2_fm}
\end{equation}
The final Stage-2 loss is
\(\mathcal{L}_{\mathrm{Stage2}}=\mathcal{L}_{\mathrm{FM}}+\lambda_{\mathrm{REPA}}\mathcal{L}_{\mathrm{REPA}}\),
where \(\mathcal{L}_{\mathrm{REPA}}\) aligns intermediate denoiser features with frozen DINOv2 features~\cite{yu2025REPA}; we use \(\lambda_{\mathrm{REPA}}=0.5\).

\vspace{-0.8em}
\subsubsection{Inference}
\vspace{-0.5em}

At inference, we initialize Gaussian noise with the normalized NIF latent shape and integrate the learned velocity field with Euler ODE sampling~\cite{Song_DDIM_2020}, using classifier-free guidance~\cite{Ho_Salimans_CFG_2022}. This produces a generated normalized latent \(\tilde{\boldsymbol{z}}_{\mathrm{gen}}\), which is denormalized as
\(\hat{\boldsymbol{z}}=\tilde{\boldsymbol{z}}_{\mathrm{gen}}\boldsymbol{\sigma}_z+\boldsymbol{\mu}_z\).
For a target size \((H',W')\), we follow the LIIF coordinate convention~\cite{chen_liif_2020}:
\begin{equation}
  \boldsymbol{q}_{i,j}
  =
  \left(
  \frac{2i+1}{H'}-1,\;
  \frac{2j+1}{W'}-1
  \right),
  \quad
  \boldsymbol{c}_{i,j}
  =
  \left(
  \frac{2}{H'},\;
  \frac{2}{W'}
  \right),
  \label{eq:coord_cell}
\end{equation}
where \(\boldsymbol{c}_{i,j}\) is the raw pixel extent from which CQAR derives \(\Delta\boldsymbol{c}_{i,j}\). Evaluating Eq.~\eqref{eq:rapd_render} with \(\hat{\boldsymbol{z}}\) on this grid yields \(\hat{\boldsymbol{x}}_{H'\times W'}\). Thus, only the query grid changes across resolutions; diffusion cost remains fixed, while rendering cost scales with queried tokens, as shown in Tab.~\ref{tab:compute_scaling}.

\section{Experiments}
\label{sec:experiments}


\vspace{-1em}
\paragraph{Dataset and Evaluation.}
Following PixelGen~\cite{ma2026pixelgen} and DeCo~\cite{Ma_DeCo_2025}, we use BLIP3o~\cite{chen2025blip3} as the primary training corpus. RaPD is pretrained on approximately 36M images and fine-tuned on 60k curated high-quality images in $512\times512$ for improved visual fidelity and text-image alignment. We evaluate reconstruction with rFID~\cite{Yao_VAVAE_2025} and text-to-image generation with GenEval~\cite{ghosh2023geneval} and DPG-Bench~\cite{hu2024ella}, following the PixelGen~\cite{ma2026pixelgen} protocol for fair comparison.

\vspace{-1em}
\paragraph{Baselines.}
RaPD is compared with representative latent- and pixel-diffusion models. Latent-diffusion models include LDM~\cite{Rombach_LDM_2022}, SD3~\cite{Esser_SD3_2024}, FLUX.1-dev~\cite{flux2024}, PixArt-$\alpha$~\cite{chen2023pixart}, and SANA~\cite{xie2025sana}. Pixel-diffusion models include PixelFlow~\cite{Chen_PixelFlow_2025}, PixNerd~\cite{Wang_PixNerd_2025}, DeCo~\cite{Ma_DeCo_2025}, and PixelGen~\cite{ma2026pixelgen}.

\vspace{-1em}
\paragraph{Implementation details.}
In Stage~1, we learn a $16\times512\times512$ NIF latent space for 10k steps on the 36M images.
Semantic distillation uses the final-layer patch tokens from a frozen DINOv2 ViT-B/14 encoder as the teacher feature $\boldsymbol{t}$.
In Stage~2, the diffusion model is pretrained for 260k steps and fine-tuned for 40k steps on $8\times$NVIDIA B200 GPUs, taking about 5 days in total.
At inference, we use $25$-step Euler ODE sampling with CFG scale $4.0$. More details can be found in Appendix.

\begin{figure}[t]
\centering
\includegraphics[width=0.98\columnwidth]{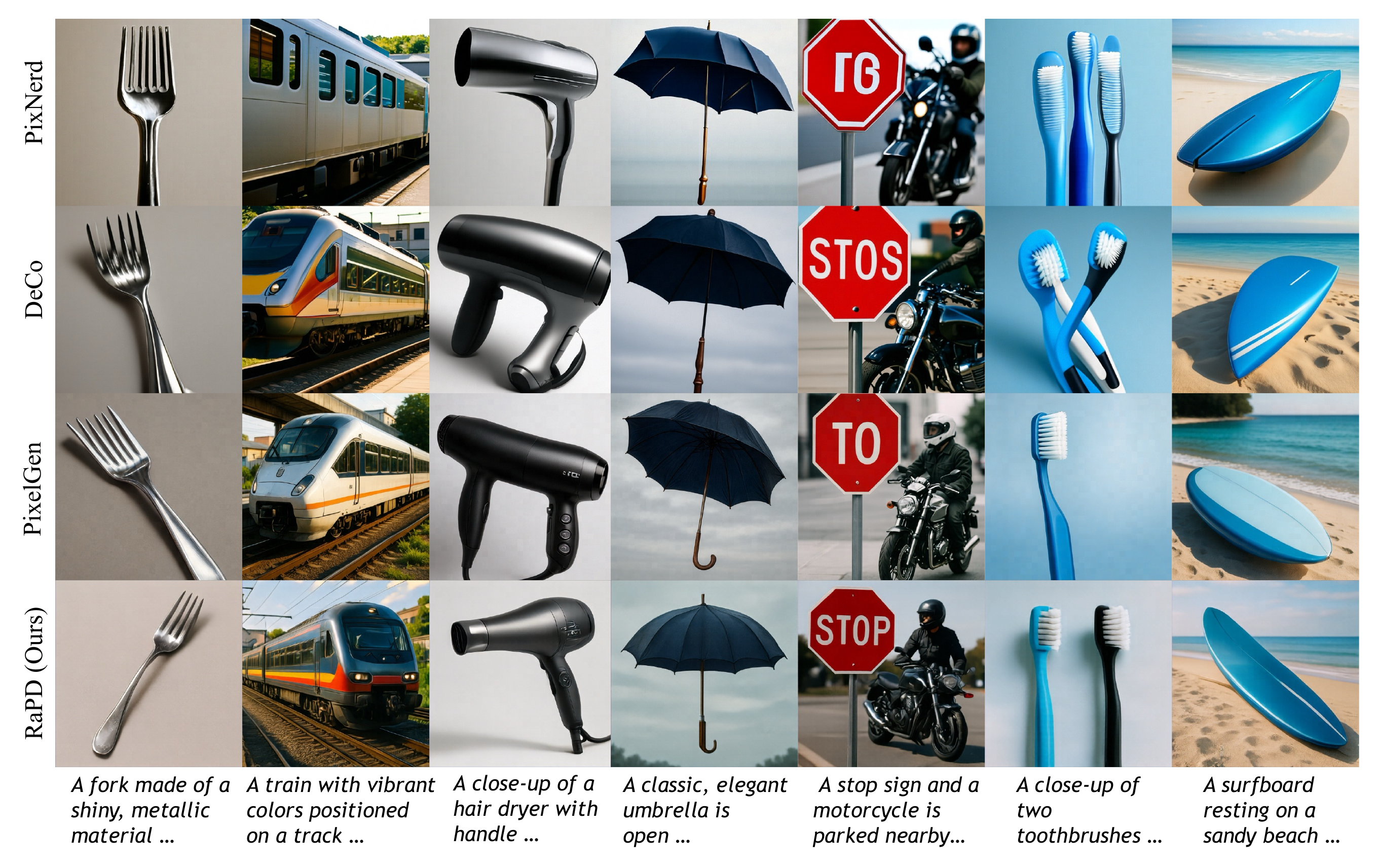}
\caption{Qualitative $512^2$ samples: RaPD vs. pixel-diffusion baselines~\cite{ma2026pixelgen, Ma_DeCo_2025, Wang_PixNerd_2025}.}
\label{fig:t2i_qualitative}
\end{figure}

\begin{table}[t]
\centering
\small
\caption{Text-to-image generation on GenEval~\cite{ghosh2023geneval} ($512\times 512$). $\ddagger$ indicates our re-runs of released models under the same seed and environment.}
\label{tab:geneval}
\setlength{\tabcolsep}{4pt}
\resizebox{0.95\linewidth}{!}{
\begin{tabular}{l|c|ccccccc}
\toprule
\multirow{2}{*}{Method} & \multirow{2}{*}{\#Params} & \multicolumn{7}{c}{GenEval $\uparrow$} \\
 &  & Single Obj. & Two Obj. & Counting & Colors & Position & Color Attr. & Overall \\
\midrule
\rowcolor{gray!15}
\multicolumn{9}{c}{$\triangledown$ \textit{Latent Diffusion Models}} \\
LDM & 1.4B & 0.92 & 0.29 & 0.23 & 0.70 & 0.02 & 0.05 & 0.37 \\
SD3 & 8B & 0.98 & 0.84 & 0.66 & 0.74 & 0.40 & 0.43 & 0.68 \\
FLUX.1-dev & 12B & 0.99 & 0.81 & 0.79 & 0.74 & 0.20 & 0.47 & 0.67 \\
SANA & 1.6B & 0.99 & 0.79 & 0.63 & 0.88 & 0.18 & 0.47 & 0.66 \\
\midrule
\rowcolor{gray!15}
\multicolumn{9}{c}{$\triangledown$ \textit{Pixel Diffusion Models}} \\
PixelFlow-XL/4 & 0.9B & -- & -- & -- & -- & -- & -- & 0.60 \\
PixNerd-XXL/16 & 1.2B & 0.97 & 0.86 & 0.44 & 0.83 & 0.71 & 0.53 & 0.73 \\
DeCo-XXL/16 & 1.1B & \textbf{1.00} & 0.92 & 0.72 & 0.91 & \textbf{0.80} & \textbf{0.79} & \textbf{0.86} \\
PixelGen-XXL/16 & 1.1B & 0.99 & 0.88 & 0.59 & 0.90 & 0.70 & 0.70 & 0.79 \\
PixNerd-XXL/16 ($\ddagger$) & 1.2B & 0.98 & 0.82 & 0.43 & 0.81 & 0.66 & 0.55 & 0.71 \\
DeCo-XXL/16 ($\ddagger$) & 1.1B & 0.99 & 0.92 & 0.65 & 0.91 & 0.71 & 0.77 & 0.83 \\
PixelGen-XXL/16 ($\ddagger$) & 1.1B & 0.99 & 0.88 & 0.58 & 0.91 & 0.70 & 0.67 & 0.79 \\
\midrule
\textbf{RaPD} & 1.1B & \underline{0.99} & \textbf{0.93} & \underline{0.72} & \textbf{0.93} & 0.77 & 0.74 & \underline{0.85} \\
\textbf{RaPD (256$\rightarrow$512)} & 1.1B & \underline{0.99} & \textbf{0.93} & \textbf{0.77} & \underline{0.92} & \underline{0.79} & \underline{0.78} & \textbf{0.86} \\
\bottomrule
\end{tabular}
}
\end{table}

\vspace{0.5cm}
\subsection{Main Results}
\label{sec:main_results}

\subsubsection{Text-to-Image Generation Compared with Baseline Methods}
\label{sec:t2i_main}

\paragraph{Qualitative Comparison.}
Fig.~\ref{fig:t2i_qualitative} compares RaPD with recent pixel-diffusion baselines at \(512\times512\). Across diverse prompts, RaPD produces more coherent global structure, cleaner local details, and fewer artifacts than competing methods. It better preserves object identity, reduces failures such as broken contours, duplicated parts, and distorted fine structures, and more reliably satisfies compositional constraints involving multiple objects, attributes, and spatial relations.

\vspace{-1em}
\paragraph{Quantitative Comparison.}
Tab.~\ref{tab:geneval} and Tab.~\ref{tab:dpg} evaluate standard text-to-image quality at $512\times512$ on GenEval~\cite{ghosh2023geneval} and DPG-Bench~\cite{hu2024ella}. RaPD (256$\rightarrow$512) follows pixel diffusion methods by pretraining at $256^2$, continued pretraining at $512^2$, and final SFT for fair comparison.
On GenEval, RaPD reaches an overall score of $0.86$, outperforming PixelGen and the reproduced DeCo results.
The improvement is most evident in compositional categories, including two-object, counting, and color, suggesting that the semantic-aligned NIF latent provides useful semantic structure for text-conditioned synthesis.
On DPG-Bench, RaPD reaches an average score of $82.9$, outperforming the reproduced pixel-diffusion baselines with strong attribute and relation scores.

\begin{figure}[t]
\centering
\includegraphics[width=\columnwidth]{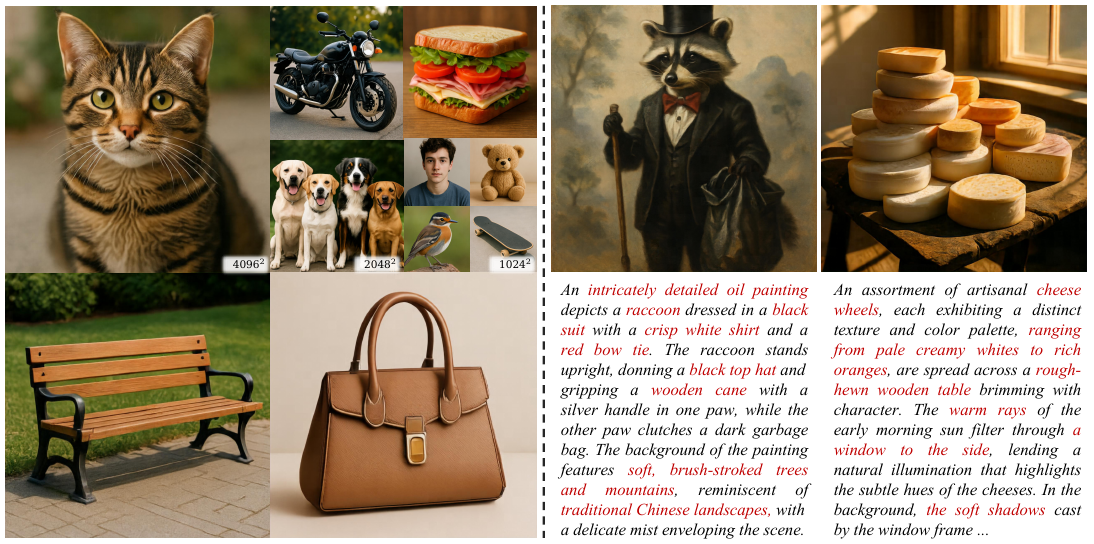}
 \caption{Arbitrary-resolution generation. Left: GenEval~\cite{ghosh2023geneval} prompts rendered at multiple output scales. Right: DPG-Bench~\cite{hu2024ella} examples showing strong long-form prompt fidelity.}
\label{fig:arbitrary_resolution}
\end{figure}

\begin{table}[t]
\centering
\small
\caption{
Text-to-image generation results on DPG-Bench~\cite{hu2024ella} ($512\times 512$). $\ddagger$ denotes results obtained by running the released models with the same random seed and environment as RaPD.
}
\label{tab:dpg}
\setlength{\tabcolsep}{3pt}
\begin{tabular}{l|c|cccccc}
\toprule
\multirow{2}{*}{Method} & \multirow{2}{*}{\#Params} & \multicolumn{6}{c}{{DPG-Bench} $\uparrow$}  \\
 &  & Global & Entity & Attribute & Relation & Other & Average \\
\midrule
\rowcolor{gray!15}
\multicolumn{8}{c}{$\triangledown$ \textit{Latent Diffusion Models}} \\
SD v2 & 0.9B & 77.7 & 78.1 & 74.9 & 80.7 & 80.7 & 68.1 \\
PixArt-$\alpha$ & 0.6B & 75.0 & 79.3 & 78.6 & 82.6 & 77.0 & 71.1 \\
SANA & 1.6B & 90.3 & 91.2 & 89.0 & 88.9 & 92.0 & 85.5 \\
\midrule
\rowcolor{gray!15}
\multicolumn{8}{c}{$\triangledown$ \textit{Pixel Diffusion Models}} \\
PixelFlow-XL/4 & 0.9B & -- & -- & -- & -- & -- & 77.9 \\
PixNerd-XXL/16 & 1.2B & 80.5 & 87.9 & 87.2 & \textbf{91.3} & 72.8 & 80.9 \\
DeCo-XXL/16 & 1.1B & -- & -- & -- & -- & -- & 81.4 \\
PixNerd-XXL/16 ($\ddagger$) & 1.2B & 87.2 & 87.1 & 88.9 & 87.1 & \textbf{89.8} & 81.5 \\
PixelGen-XXL/16 ($\ddagger$) & 1.1B & 81.4 & 88.8 & 86.7 & 88.9 & 88.9 & 80.4 \\
DeCo-XXL/16 ($\ddagger$) & 1.1B & \textbf{88.6} & 88.4 & 88.9 & 87.6 & 86.0 & 81.2 \\
\midrule
\textbf{RaPD} & 1.1B & 85.1 & 87.7 & \underline{88.9} & \underline{89.6} & 84.1 & \underline{81.6} \\
\textbf{RaPD (256$\rightarrow$512)} & 1.1B & 81.3 & \textbf{89.9} & \textbf{89.7} & 88.9 & 81.2 & \textbf{82.9} \\
\bottomrule
\end{tabular}
\vspace{-2em}
\end{table}

\vspace{-0.2em}

\subsubsection{Evaluation of Arbitrary-Resolution Generation.}
\label{sec:arbitrary_resolution}

\vspace{-0.5em}
\paragraph{Qualitative and Quantitative Results.}
Fig.~\ref{fig:arbitrary_resolution} shows RaPD results at $1024^2$, $2048^2$, and $4096^2$ on GenEval and DPG-Bench examples.
Although the NIF latent is learned with limited multi-resolution supervision up to $512^2$, RaPD generalizes well to out-of-distribution higher resolutions.
The high-resolution samples remain coherent and contain clear boundaries and rich local details.
The DPG-Bench results further indicate strong fidelity to long and complex prompts at large output scales.

\begin{table}[t]
\centering
\caption{Comparison with high-resolution generation baselines.}
\label{tab:comparison}
\setlength{\tabcolsep}{3pt}
\begin{tabular}{
m{0.24\linewidth}
|>{\centering\arraybackslash}m{0.11\linewidth}
|>{\centering\arraybackslash}m{0.12\linewidth}
 >{\centering\arraybackslash}m{0.12\linewidth}
|>{\centering\arraybackslash}m{0.12\linewidth}
 >{\centering\arraybackslash}m{0.12\linewidth}
}
\toprule
\multirow{2}{*}{Method}
& \multirow{2}{*}{\#Params}
& \multicolumn{2}{c|}{$1024^2$}
& \multicolumn{2}{c}{$2048^2$} \\
& &
GenEval $\uparrow$ & DPG $\uparrow$
& GenEval $\uparrow$ & DPG $\uparrow$ \\
\midrule
SANA
& 1.6B
& 0.66 & \textbf{84.8}
& 0.70 & 83.1 \\
SANA-v1.5
& 1.6B
& 0.82 & 84.5
& 0.80 & \textbf{83.8} \\
FLUX.1-dev
& 12B
& 0.68 & 83.1
& 0.66 & 82.8 \\
\textbf{RaPD (256$\rightarrow$1024)}
& {1.1B}
& \textbf{0.85} & 80.3
& \textbf{0.85} & 80.1 \\
\bottomrule
\end{tabular}
\end{table}

\paragraph{Quantitative Comparison with High-Resolution Methods}
For a fair comparison, we further SFT the $512$-resolution RaPD model
to $1024$ resolution. As shown in Table~\ref{tab:comparison}, RaPD
achieves the best GenEval score of $0.85$ at both $1024$ and $2048$
resolutions. Its lower DPG scores suggest that fine-grained attributes
and relations remain more challenging, leaving room for improvement in
detailed prompt adherence.

\paragraph{Time and Computational Cost.}
Tab.~\ref{tab:compute_scaling} shows that RaPD scales much more favorably than fixed-grid pixel diffusion at high resolutions.
RaPD keeps the expensive denoising stage nearly fixed, while PixelGen and DeCo grow quadratically with the output resolution, leading to a large increase in inference time.
At $4096^2$, RaPD requires only $2.13$s, whereas PixelGen and DeCo both exceed $61$s.

\begin{figure}[t]
\centering
\includegraphics[width=\columnwidth]{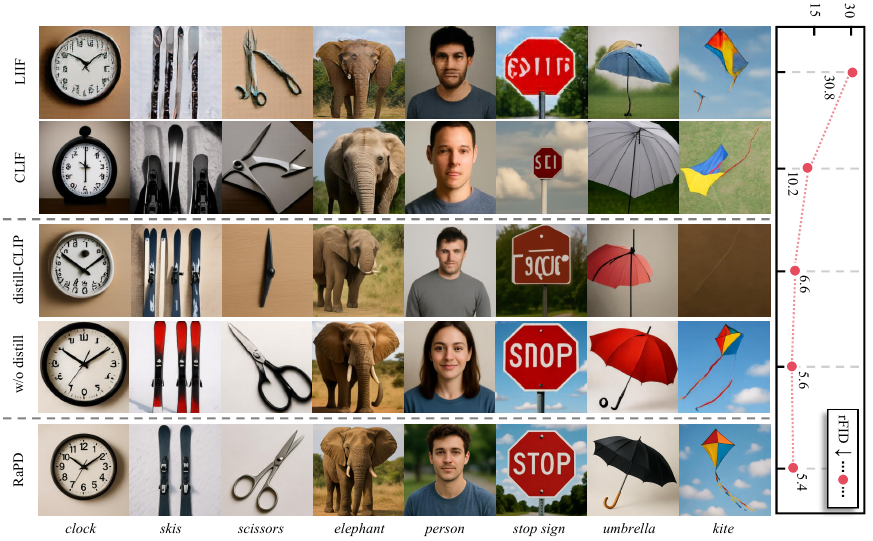}
\caption{Ablations of Semantic Distillation and CQAR.}
\label{fig:ablation}
\end{figure}

\begin{table}[t]
\centering
\caption{
Per-image inference time from $1\times$ to $8\times$ output scales on a B200 GPU. Values in parentheses indicate the increase over $512\times512$.
}
\label{tab:compute_scaling}
\setlength{\tabcolsep}{4pt}
\begin{tabular}{l|cc|cc|cc|cc}
\toprule
\multirow{2}{*}{Method} &
\multicolumn{2}{c|}{$512\times512$ ({1$\times$})} &
\multicolumn{2}{c|}{$1024\times1024$ ({2$\times$})} &
\multicolumn{2}{c|}{$2048\times2048$ ({4$\times$})} &
\multicolumn{2}{c}{$4096\times4096$ ({8$\times$})} \\
&
\#Tokens & Time (s) &
\#Tokens & Time (s) &
\#Tokens & Time (s) &
\#Tokens & Time (s) \\
\midrule
PixelGen
& 1,024 & 0.19
& 4,096 & 0.60
& 16,384 & 4.72
& 65,536 & 61.46 \textcolor{red}{(+61.27)} \\
DeCo
& 1,024 & 0.21
& 4,096 & 0.67
& 16,384 & 5.03
& 65,536 & 62.72 \textcolor{red}{(+62.51)} \\
\midrule
\textbf{RaPD}
& 1,024 & 0.23
& 1,024 & 0.34
& 1,024 & 0.68
& 1,024 & 2.13 \textcolor{Green}{(+1.9)} \\
\bottomrule
\end{tabular}
\end{table}

\begin{table}[t]
\centering
\caption{Ablation on Semantic Distillation and REPA.}
\label{tab:ablation}
\setlength{\tabcolsep}{3pt}
\begin{tabular}{l|cc}
\toprule
Setting & GenEval $\uparrow$ & DPG $\uparrow$ \\
\midrule
w/o Stage-1 Semantic Distillation & 0.83 & 78.9 \\
w/o Stage-2 REPA & 0.61 & 56.6 \\
RaPD (Full model) & \textbf{0.85} & \textbf{81.3} \\
\bottomrule
\end{tabular}
\end{table}

\subsection{Model Analysis}
\label{sec:analysis_experiments}

\paragraph{Ablation on Semantic Distillation and REPA.}
Fig.~\ref{fig:ablation} evaluates Semantic Distillation. Without it (\textit{w/o distill}), generation exhibits weaker object structure and less stable semantic layouts, showing that reconstruction supervision alone is insufficient for NIF-latent text-to-image generation. Replacing DINOv2 with CLIP (\textit{distill-CLIP}) further degrades semantic quality, supporting DINOv2 as a stronger teacher for generation-aware NIF latents. The quantitative results without Semantic Distillation are reported in Table~\ref{tab:ablation}.
REPA has also been validated by many recent generative modeling
studies for improving convergence and generation quality. The two
objectives are complementary rather than redundant: Semantic
Distillation shapes the semantic structure of the NIF latent, while
REPA regularizes the diffusion denoiser.

\begin{figure}[!h]
\centering
\begin{minipage}[t]{0.65\columnwidth}
    \centering
    \includegraphics[width=0.9\linewidth]{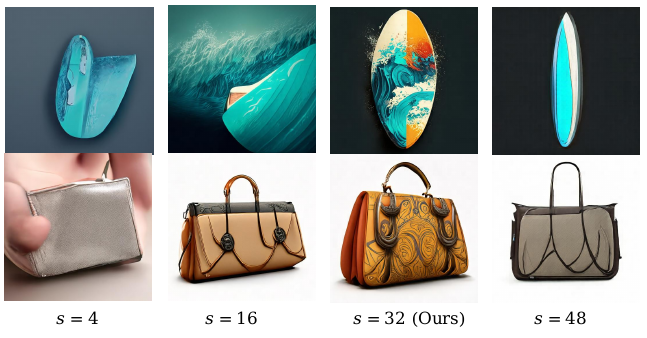}
    \vspace{-0.4cm}
    \caption{Generation under different timestep shift (prompt: ``\textit{a photo of a surfboard / handbag }'').}
    \label{fig:shift}
\end{minipage}
\hfill
\begin{minipage}[t]{0.31\columnwidth}
    \centering
    \includegraphics[width=0.9\linewidth]{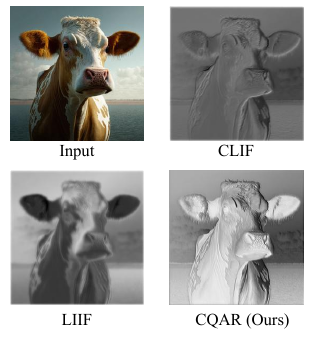}
    \vspace{-0.4cm}
    \caption{Latent feature visualization.}
    \label{fig:feature}
\end{minipage}
\end{figure}

\paragraph{Exploration of Latent-Density-Aware Timestep Shift.}
Fig.~\ref{fig:shift} studies the Latent-Density-Aware Timestep Shift in learning our dense NIF latent.
Small shift values lead to weak object formation and poor semantic organization, showing that schedules suited to compressed VAE latents do not transfer well to our full-resolution NIF representation.
Increasing the shift substantially improves generation quality, and $s=32$ provides a better trade-off between clear object layout and rich visual detail.
When the shift becomes too large ($s=48$), the generated images become overly simplified.

\paragraph{The Effect of CQAR.}
Fig.~\ref{fig:nif_compare} compares CQAR with LIIF- and CLIF-style renderers on image reconstruction.
CQAR produces sharper reconstructions with clearer boundaries and better recovery of fine structures. Following DC-AE 1.5~\cite{chen2025dc}, Fig.~\ref{fig:feature} visualizes the learned NIF features and shows that CQAR preserves clearer spatial details in the feature maps.
Fig.~\ref{fig:ablation} further compares the generation results. RaPD achieves the lowest rFID and the best visual quality, indicating that CQAR effectively alleviates the reconstruction--generation gap of continuous NIF representations.

\section{Conclusion and Limitations}
\label{sec:conclusion}

In this paper, we presented \textbf{RaPD}, which performs diffusion in a continuous Neural Image Field latent and decodes it through a coordinate-queried windowed-attention renderer, achieving resolution-agnostic image generation while keeping the diffusion cost fixed regardless of the output size. Three directions naturally extend this work: \emph{(i)} scaling to larger datasets and models; \emph{(ii)} extending the continuous NIF latent to controllable generation and editing; and \emph{(iii)} hardware-aware inference acceleration to fully realize RaPD's efficiency potential at high resolutions.

\vspace{-1em}
   \paragraph{Limitations.}
RaPD still has several limitations. 
\emph{(1)} It is trained on approximately 36 million BLIP3o images, substantially fewer than the data used by commercial-scale models such as SD3 and FLUX, which limits its absolute generation quality. 
\emph{(2)} The frozen Qwen3-1.7B text encoder is relatively compact and may constrain complex prompt understanding compared with systems paired with larger language models. 
\emph{(3)} The current inference stack does not yet incorporate FlashAttention~\cite{dao2022flashattentionfastmemoryefficientexact}, fused window-attention kernels for CQAR, FP8 quantization, or compiler-level fusion; therefore, the reported latency and memory should be viewed as conservative upper bounds.



\section*{Acknowledgment}
This work was supported in part by the National Natural Science Foundation of China under Grant No.  62473290, the the National Key R\&D Program of China under Grant No. 2024YFB3311801, the Shanghai Municipal Science and Technology Commission under Grant No. 25511102800, the Shanghai Municipal Commission of Economy and Information Technology under Grant No. 2024-GZL-RGZN-01008.

\nocite{*}
\bibliographystyle{plainnat}
\bibliography{main}

\clearpage
\appendix

\noindent{\LARGE\bfseries Appendix}\par\medskip

\noindent This supplementary material includes the following:
\begin{itemize}[leftmargin=14pt, labelsep=0.5em, itemindent=0pt, label={}]
  \item[A.] Related Work
  \item[B.] More Implementation Details
  \item[C.] Additional Experimental Results
\end{itemize}

\section{Related Work}

\begin{table}[b]
\centering
\caption{Comparison of representative image generation method categories. \textbf{Pixel-Space}: the forward/reverse diffusion process operates in pixel space without lossy VAE compression. \textbf{Arb.\ Resolution Output}: supports synthesizing images at arbitrary target resolutions at inference time. \textbf{Continuous Semantic Prior}: the generative prior models spatial continuity with semantically-aware features rather than treating pixels as discrete grid elements. \textbf{Decoder Type}: the decoding mechanism used to produce pixel values.
Per-category methods: Latent Diffusion Models~\cite{Rombach_LDM_2022,Esser_SD3_2024,flux2024}; Latent Diffusion w/ Flexible Resolution~\cite{lu2024fit,Wang_FiTv2_2024,Gao_ArbitraryScale_2023,Chen_INFD_2024,InfGen_2025,Wang_Bai_Yue_Ouyang_Zhang_2025}; Pixel Diffusion w/ Fixed Resolution~\cite{Yu_PixelDiT_2025,Chen_DiP_2025,Li_He_JiT_2026,ma2026pixelgen,Ma_DeCo_2025,Crowson_HDiT_2024,EPG_NoVAE_2026}; Pixel Diffusion w/ Partial Continuity~\cite{Wang_PixNerd_2025}.
\checkmark~= fully supported; $\circ$~= partially supported; \texttimes~= not supported.}
\label{tab:method_comparison}
\resizebox{\linewidth}{!}{%
\begin{tabular}{lcccc}
\toprule
\textbf{Method Category} & \textbf{Pixel-Space} & \textbf{Arb.\ Resolution Output} & \textbf{Continuous Semantic Prior} & \textbf{Decoder Type} \\
\midrule
Latent Diffusion Models
  & \texttimes & \texttimes & \texttimes & Fixed Conv \\
Latent Diffusion w/ Flexible Resolution
  & \texttimes & $\circ$ & \texttimes & INR / Conv \\
Pixel Diffusion w/ Fixed Resolution
  & \checkmark & \texttimes & \texttimes & None \\
Pixel Diffusion w/ Partial Continuity
  & \checkmark & $\circ$ & $\circ$ & Partial INR \\
\midrule
\textbf{Ours (RaPD)}
  & \checkmark & \checkmark & \checkmark & CQAR \\
\bottomrule
\end{tabular}%
}
\end{table}

\subsection{Image Generation and Resolution Flexibility}

Image generative models have evolved from GANs~\cite{goodfellow_GAN_2014, karras2018_PGGAN, karras2019StyleGAN, karras2020StyleGAN2} through denoising diffusion probabilistic models~\cite{sohldickstein2015DiffusionModel, Ho_DDPM_2020, Song_DDIM_2020, song2021ScoreMatching} to the Diffusion Transformer (DiT) paradigm~\cite{Peebles_Xie_2022, Ma_Goldstein_Albergo_Boffi_Vanden, Gao_Zhou_Cheng_Yan_2024}, which has become the dominant backbone for high-quality image synthesis.
Latent Diffusion Models (LDMs)~\cite{Rombach_LDM_2022, Podell_SDXL_2023, Esser_SD3_2024, flux2024} dramatically reduce computational cost by operating in a compressed VAE latent space, but this compression introduces an information bottleneck that degrades fine-grained spatial details and fixes the output resolution~\cite{Kingma_VAE_2014, Chen_DCAE_2025, Yao_VAVAE_2025, Leng_REPA-E_2025}.
Motivated by this, pixel diffusion methods~\cite{Yu_PixelDiT_2025, Chen_DiP_2025, Li_He_JiT_2026, Ma_DeCo_2025, Chen_PixelFlow_2025, ma2026pixelgen, Crowson_HDiT_2024, EPG_NoVAE_2026, Hoogeboom_Heek_Salimans_2023} bypass the VAE entirely to preserve complete spatial information, with recent work demonstrating competitive quality at resolutions up to 1024$\times$1024 in pixel space.
Tab.~\ref{tab:method_comparison} summarizes the key differences among these method categories.

Achieving resolution flexibility within these frameworks remains challenging. Within the LDM paradigm, several methods pair the diffusion process with flexible decoders to support variable output resolutions~\cite{lu2024fit, Wang_FiTv2_2024, Gao_ArbitraryScale_2023, Chen_INFD_2024, InfGen_2025, Wang_Bai_Yue_Ouyang_Zhang_2025, Wang_Li_Song_Li_Ge_Zheng_Wang_2024}; however, the diffusion process itself still operates on a fixed-resolution compressed latent, confining the model's semantic planning to a single scale. Early GAN-based approaches replace spatial convolutions with coordinate-queried MLPs for multi-scale synthesis~\cite{Anokhin_CIPS_2021, INRGAN2021, Chai_ArbitraryScale_2022}, but suffer from training instability and degraded quality at out-of-distribution resolutions~\cite{FID_2017}. Among pixel-space methods, PixNerd~\cite{Wang_PixNerd_2025} is closest to resolution-agnostic generation, rendering beyond the training grid via a patch-wise neural field decoder, but its backbone transformer still processes a fixed token count at the pretrained resolution, restricting flexibility to the decoding stage alone. True arbitrary-resolution generation requires the entire pipeline to operate on representations decoupled from any fixed spatial grid.

\subsection{Implicit Neural Representations for Generation}

Implicit neural representations (INRs)~\cite{Sitzmann_INR_2020, mildenhall2020NERF} model signals as continuous functions mapping coordinates to values. In the 2D image domain, LIIF~\cite{chen_liif_2020} and MetaSR~\cite{MetaSR_2019} established the paradigm of decoding a shared feature map at arbitrary resolutions via coordinate-conditioned MLPs. Subsequent work applied this paradigm to generation: IDM~\cite{Chen_IDM_2023} performs INR decoding at every denoising step but at prohibitive cost; INFD~\cite{Chen_INFD_2024} applies diffusion to INR-based latents using a convolutional rendering module (CLIF); and InfGen~\cite{InfGen_2025} replaces the VAE decoder with a transformer-based generator for single-step extrapolation to 4K.

However, all existing INR-based representations are still optimized for reconstruction, not generation, suffering from two intertwined limitations. First, their latent features, trained with pixel-level losses as listed in Eq.~\ref{eq:nif-trainloss}, encode low-level texture statistics rather than the semantic structure required by a generative model. Second, their decoders (pointwise MLPs~\cite{chen_liif_2020} or shallow CNNs~\cite{Chen_INFD_2024}) lack the receptive field to exploit semantic features for coherent synthesis. Although recent works on foundation-model alignment, such as REPA~\cite{yu2025REPA, Wu_Zhang_Shi_Gao_Chen_Wang_Chen_Gao_Tang_Yang_2025, Leng_REPA-E_2025} and VA-VAE~\cite{Yao_VAVAE_2025} leverage this principle to accelerate diffusion training and balance VAE capacity respectively, our reconstruction--generation gap is fundamentally different: it concerns the semantic content of the NIF latent itself, and arises even when latent capacity is unbounded. Whereas REPA aligns denoiser features and VA-VAE shapes a separate VAE post-hoc, our \emph{Semantic Distillation} directly shapes the NIF latent itself during its very first training stage, making generation-aware structure available to the diffusion process from the start.

RaPD addresses both limitations jointly: Semantic Representation Guidance (SRG) enriches the latent with semantic structure through a foundation-model distillation procedure, and a Coordinate-Queried Attention Renderer (CQAR) replaces local decoders with coordinate-conditioned, scale-aware Transformer blocks~\cite{liu2021swin_transformer} for inter-pixel semantic reasoning at arbitrary resolutions.

\section{More Implementation Details}
\subsection{Pseudocode of RaPD}
\label{app:pseudocode}

We summarize the training pipeline (Stage-1 + Stage-2) of RaPD in Alg.~\ref{alg:rapd_training} and the arbitrary-resolution inference pipeline in Alg.~\ref{alg:rapd_inference}.

\begin{algorithm}[h]
\small
\caption{RaPD: training pipeline (Stage-1 + Stage-2).}
\label{alg:rapd_training}
\begin{algorithmic}[1]
\Statex \textbf{Stage-1: NIF Semantic Distillation}
\State Sample low-/high-resolution image pair $\boldsymbol{x}_{\mathrm{lr}}, \boldsymbol{x}_{\mathrm{hr}}$ with scale $r\!\sim\!\mathcal{U}[1,2]$
\State $\boldsymbol{z} \gets E(\boldsymbol{x}_{\mathrm{lr}})$ \Comment{encode to NIF latent at full spatial resolution}
\State $\hat{\boldsymbol{x}}_{\mathrm{hr}} \gets G(\boldsymbol{z}, \boldsymbol{q}_{\mathrm{hr}}, \Delta\boldsymbol{c}_{\mathrm{hr}})$ \Comment{render the HR patch via CQAR}
\State $\boldsymbol{z}_{\mathrm{proj}} \gets \mathrm{Proj}(\mathrm{Pool}(\boldsymbol{z}))$ \Comment{project $\boldsymbol{z}$ to the DINOv2 patch space}
\State $\mathcal{L}_{\mathrm{Stage1}} \gets \mathcal{L}_{\mathrm{rec}} + w_{\mathrm{adapt}}\!\cdot\!\mathcal{L}_{\mathrm{distill}}$ \Comment{reconstruction + semantic distillation}
\State Update $E, G$ by $\nabla \mathcal{L}_{\mathrm{Stage1}}$;\ after training, compute per-channel $\boldsymbol{\mu}_z, \boldsymbol{\sigma}_z$ on the training set
\Statex
\Statex \textbf{Stage-2: Flow Matching with Generation-aware NIF}
\State $\tilde{\boldsymbol{z}} \gets (E(\boldsymbol{x}) - \boldsymbol{\mu}_z) / \boldsymbol{\sigma}_z$ \Comment{normalized NIF latent serves as the data endpoint $\boldsymbol{x}_1$}
\State $\xi\!\sim\!\mathcal{N}(0,1)$,\ $t \gets \mathrm{shift}_s(\mathrm{sigmoid}(\xi))$ \Comment{logit-normal $t$ with the SRG-prescribed shift $s$}
\State $\boldsymbol{\epsilon}\!\sim\!\mathcal{N}(\boldsymbol{0}, \boldsymbol{I})$;\ \ $\boldsymbol{x}_t \gets (1-t)\,\boldsymbol{\epsilon} + t\,\tilde{\boldsymbol{z}}$ \Comment{noised latent at timestep $t$}
\State $\tilde{\boldsymbol{z}}_\theta \gets D_\theta(\boldsymbol{x}_t, t, \boldsymbol{y}_{\mathrm{cond}})$ \Comment{denoiser predicts the clean normalized latent}
\State $\mathcal{L}_{\mathrm{Stage2}} \gets \mathcal{L}_{\mathrm{FM}} + \lambda_{\mathrm{REPA}}\!\cdot\!\mathcal{L}_{\mathrm{REPA}}$ \Comment{flow matching + REPA feature alignment}
\State Update $\theta$ by $\nabla \mathcal{L}_{\mathrm{Stage2}}$
\end{algorithmic}
\end{algorithm}

\begin{algorithm}[h]
\small
\caption{RaPD: arbitrary-resolution inference.}
\label{alg:rapd_inference}
\begin{algorithmic}[1]
\State \textbf{Input:} text prompt $\boldsymbol{y}$ and target output size $(H', W')$
\State $\boldsymbol{\epsilon}\!\sim\!\mathcal{N}(\boldsymbol{0}, \boldsymbol{I})$ with the shape of $\tilde{\boldsymbol{z}}$ \Comment{initialize Gaussian noise at the NIF latent shape}
\State $\tilde{\boldsymbol{z}}_\theta \gets \mathrm{EulerODE}(D_\theta, \boldsymbol{\epsilon}, \boldsymbol{y}_{\mathrm{cond}})$ \Comment{denoise once with CFG; cost independent of $(H', W')$}
\State $\hat{\boldsymbol{z}} \gets \tilde{\boldsymbol{z}}_\theta \cdot \boldsymbol{\sigma}_z + \boldsymbol{\mu}_z$ \Comment{denormalize back to the encoder latent scale}
\State $(\boldsymbol{q}, \boldsymbol{c}) \gets \mathrm{CoordGrid}(H', W')$ \Comment{construct query coordinates and cell sizes for the target}
\State $\hat{\boldsymbol{x}}_{H'\!\times\!W'} \gets G(\hat{\boldsymbol{z}}, \boldsymbol{q}, \boldsymbol{c})$ \Comment{a single CQAR pass renders the final image}
\State \Return $\hat{\boldsymbol{x}}_{H'\!\times\!W'}$
\end{algorithmic}
\end{algorithm}

\subsection{Continuous Coordinate System}
\label{app:coord}

Following LIIF~\cite{chen_liif_2020}, the spatial domain is normalized to $[-1,1]^2$.
For a target output resolution $(H', W')$, we generate:
\begin{itemize}
  \item A coordinate grid $\mathbf{q}\!\in\!\mathbb{R}^{H'\times W'\times 2}$, where $\mathbf{q}_{i,j}$ gives the center of output pixel $(i,j)$ in normalized coordinates.
  \item A cell-size grid $\mathbf{c}\!\in\!\mathbb{R}^{H'\times W'\times 2}$ with $\mathbf{c}_{i,j} = (2/H', 2/W')$, encoding the spatial extent of each output pixel as an explicit scale indicator. Smaller cells correspond to higher output resolutions, providing the renderer with resolution awareness.
\end{itemize}

\subsection{Latent Normalization}
\label{app:normalization}

After Stage~1 training, per-channel mean $\boldsymbol{\mu}_z\!\in\!\mathbb{R}^C$ and standard deviation $\boldsymbol{\sigma}_z\!\in\!\mathbb{R}^C$ are computed over the training dataset.
The normalized latent is:
\begin{equation}
  \tilde{\mathbf{z}} = \frac{\mathbf{z} - \boldsymbol{\mu}_z}{\max(\boldsymbol{\sigma}_z, 10^{-6})},
\end{equation}
which approximately follows $\mathcal{N}(\boldsymbol{0},\boldsymbol{I})$, ensuring well-conditioned inputs for the diffusion model.
The inverse transformation $\mathbf{z} = \tilde{\mathbf{z}} \cdot \boldsymbol{\sigma}_z + \boldsymbol{\mu}_z$ is applied before CQAR rendering.

\subsection{CQAR Architecture Details}
\label{app:cqar}

\paragraph{Window Transformer Blocks.}
Each of the 4 blocks applies:
\begin{align}
  \mathbf{x}' &= \mathbf{x} + \mathrm{W-MSA}\!\bigl(\mathrm{LN}(\mathbf{x})\bigr), \\
  \mathbf{x}'' &= \mathbf{x}' + \mathrm{GatedFFN}\!\bigl(\mathrm{LN}(\mathbf{x}')\bigr),
\end{align}
where $\mathrm{W-MSA}$ is windowed multi-head self-attention with window size $W\!=\!8$ and $N_h\!=\!4$ heads.
Odd-indexed blocks use shifted windows (shift $W/2$) with an attention mask for cross-window information exchange.

\paragraph{Windowed Self-Attention with Relative Position Bias.}
Within each $W\!\times\!W$ window:
\begin{equation}
  \mathrm{Attn}(\mathbf{Q}, \mathbf{K}, \mathbf{V}) = \mathrm{softmax}\!\left(\frac{\mathbf{Q}\mathbf{K}^\top}{\sqrt{d_h}} + \mathbf{B}\right)\mathbf{V},
\end{equation}
where $d_h = D/N_h$ is the head dimension and $\mathbf{B}\!\in\!\mathbb{R}^{W^2\times W^2}$ is a learned relative position bias parameterized by a table of size $(2W\!-\!1)^2\!\times\!N_h$.

\paragraph{Shifted-Window Self-Attention (SW-MSA).}
To enable cross-window information exchange without breaking the linear cost of windowed attention, every odd-indexed block uses Swin-style shifted-window attention~\cite{liu2021swin_transformer}. Concretely, the token grid is first cyclically shifted by $(W/2, W/2)$, partitioned into the same $W\!\times\!W$ windows, and processed by the windowed attention in the equation above. Tokens that originate from different real windows but are merged into the same shifted window are blocked from attending to each other through an additive attention mask:
\begin{equation}
  \mathbf{M}_{ij} =
  \begin{cases}
    0, & \text{if tokens } i, j \text{ belong to the same real window},\\
    -\infty, & \text{otherwise},
  \end{cases}
\end{equation}
which is added to the pre-softmax logits. After attention, the cyclic shift is reversed so that the spatial layout is preserved. This shifted--regular alternation propagates information across window boundaries across consecutive blocks while keeping the per-block attention cost at $O(W^2)$ per token.

\paragraph{Gated Feed-Forward Network.}
\begin{equation}
  \mathrm{GatedFFN}(\mathbf{x}) = \mathrm{Linear}\!\bigl(\mathrm{GELU}(\mathbf{x}_1) \odot \mathbf{x}_2\bigr),
  \quad [\mathbf{x}_1, \mathbf{x}_2] = \mathrm{Linear}(\mathbf{x}),
\end{equation}
with expansion factor $2\times$.
The entire renderer is convolution-free, consisting exclusively of linear layers, layer normalization, and attention.
A final LayerNorm and linear projection ($D\!\to\!3$) produce the RGB output $\hat{\mathbf{x}}\!\in\!\mathbb{R}^{3\times H'\times W'}$.

\subsection{PixGen-style DiT Denoiser Architecture}
\label{app:denoiser}

We adopt a PixGen-style DiT denoiser~\cite{Peebles_Xie_2022, ma2026pixelgen}:
\begin{enumerate}
  \item \textbf{Patch embedding}: unfolds $\tilde{\mathbf{z}}\!\in\!\mathbb{R}^{16\times 512\times 512}$ into $16\!\times\!16$ patches, yielding 1024 image tokens compressed through a bottleneck ($4096\!\to\!1024\!\to\!1536$).
  \item \textbf{Text encoder}: frozen Qwen3-1.7B~\cite{Yang_Li_Yang_Zhang_Hui_Zheng_Yu_Gao_Huang_Lv_2025} produces 128 tokens at dimension 2048, projected to hidden dimension 1536.
  \item \textbf{Text-refinement blocks}: 4 blocks of self-attention on text tokens with AdaLN modulation from the timestep embedding.
  \item \textbf{Image denoising blocks}: 16 DiT blocks~\cite{Peebles_Xie_2022} with cross-attention between image queries and text keys/values, using RoPE on image tokens, RMSNorm on Q/K, SwiGLU FFN, and gated attention. Both attention and FFN are modulated by AdaLN from the timestep condition.
  \item \textbf{Final layer}: LayerNorm followed by linear projection, folding output back to $\tilde{\mathbf{z}}_{\theta}\!\in\!\mathbb{R}^{16\times 512\times 512}$.
\end{enumerate}
The denoiser uses hidden dimension 1536 with 24 attention heads.
A REPA loss~\cite{yu2025REPA} (weight 0.5) aligns the denoiser image-token features at transformer block index 8 with the final-layer patch-token features of a frozen DINOv2 ViT-B/14 encoder.

Timesteps follow the logit-normal distribution $t = \mathrm{sigmoid}(\xi),\; \xi \sim \mathcal{N}(0,\,1)$, shifted via $t' = t / (t + (1-t) \cdot s)$ with $s\!=\!32$ and $\delta_2\!=\!10^{-8}$.

\subsection{Other Details}
\label{app:impl}

Table~\ref{tab:model_spec} summarizes all hyperparameters.

\begin{table}[h]
\centering
\caption{RaPD model specifications and training hyperparameters.}
\label{tab:model_spec}
\small
\begin{tabular}{@{}lll@{}}
\toprule
\textbf{Component} & \textbf{Parameter} & \textbf{Value} \\
\midrule
\multicolumn{3}{@{}l}{\textit{Stage 1: NIF AutoEncoder}} \\
\quad Encoder (EDSR) & Channels / ResBlocks & 16 / 8 \\
\quad CQAR Renderer & Hidden / Blocks / Heads / Window & 256 / 4 / 4 / 8 \\
\quad Distill Teacher & Model & DINOv2 ViT-B/14 (frozen) \\
\quad Distill Margins & $m_{\mathrm{cos}}$ / $m_{\mathrm{dist}}$ / $w_{\mathrm{base}}$ & 0.5 / 0.25 / 0.1 \\
\quad Training & Steps / LR / Batch / Scale & 10K / $2\!\times\!10^{-4}$ / 8 / $[1,2]$ \\
\midrule
\multicolumn{3}{@{}l}{\textit{Stage 2: Flow Matching Denoiser}} \\
\quad DiT Denoiser & Hidden / Blocks / Heads & 1536 / 16 / 24 \\
\quad & Patch size / Bottleneck & 16 / 1024 \\
\quad Text Encoder & Qwen3-1.7B (frozen) & 128 tokens, dim 2048 \\
\quad Flow Matching & Timestep shift $s$ & \textbf{32.0} \\
\quad & $t$ sampling & LogitNormal ($\mu\!=\!0.0$, $\sigma\!=\!1.0$) \\
\quad & $\delta_2$ & $10^{-8}$ \\
\quad & REPA weight & 0.5 \\
\quad Training & Steps / LR / EMA decay & 300K / $2\!\times\!10^{-4}$ / 0.9999 \\
\quad & Precision / Grad clip / Batch & bfloat16 / 1.0 / 256 \\
\midrule
\multicolumn{3}{@{}l}{\textit{Inference}} \\
\quad Sampler & Euler ODE / 25 steps & CFG $w\!=\!4.0$ \\
\quad Resolution & Arbitrary & Tested 512--4096 ($1\!\times\!$--$8\!\times\!$) \\
\bottomrule
\end{tabular}
\end{table}

\paragraph{Datasets.}
Stage~1 uses high-resolution images from BLIP3o for autoencoder training with multi-resolution supervision.
Stage~2 uses BLIP3o image--text pairs (long and short caption variants) streamed via WebDataset at $512\!\times\!512$ resolution.
We evaluate on GenEval~\cite{ghosh2023geneval} and DPG-Bench~\cite{hu2024ella}.

\paragraph{Multi-Resolution Training.}
For each Stage~1 sample: (1)~sample scale $r \sim \mathcal{U}[1, 2]$; (2)~extract a high-resolution crop of size $(\lfloor r H_0 \rceil, \lfloor r W_0 \rceil)$ as $\boldsymbol{x}_{\mathrm{tar}}$, where $(H_0, W_0)$ is the fixed encoder input resolution; (3)~downsample $\boldsymbol{x}_{\mathrm{tar}}$ to obtain encoder input $\boldsymbol{x}_{\mathrm{in}}$; (4)~construct the full target coordinate grid; (5)~supervise the CQAR output against every pixel of $\boldsymbol{x}_{\mathrm{tar}}$ via the standard NIF reconstruction loss ($\ell_1$ + LPIPS).

\section{Additional Experimental Results}
\label{app:add_exp}

\subsection{Additional Qualitative Comparisons}
\label{app:add_qualitative}

\begin{figure}[h!]
  \makebox[\linewidth][c]{\includegraphics[width=1.2\linewidth]{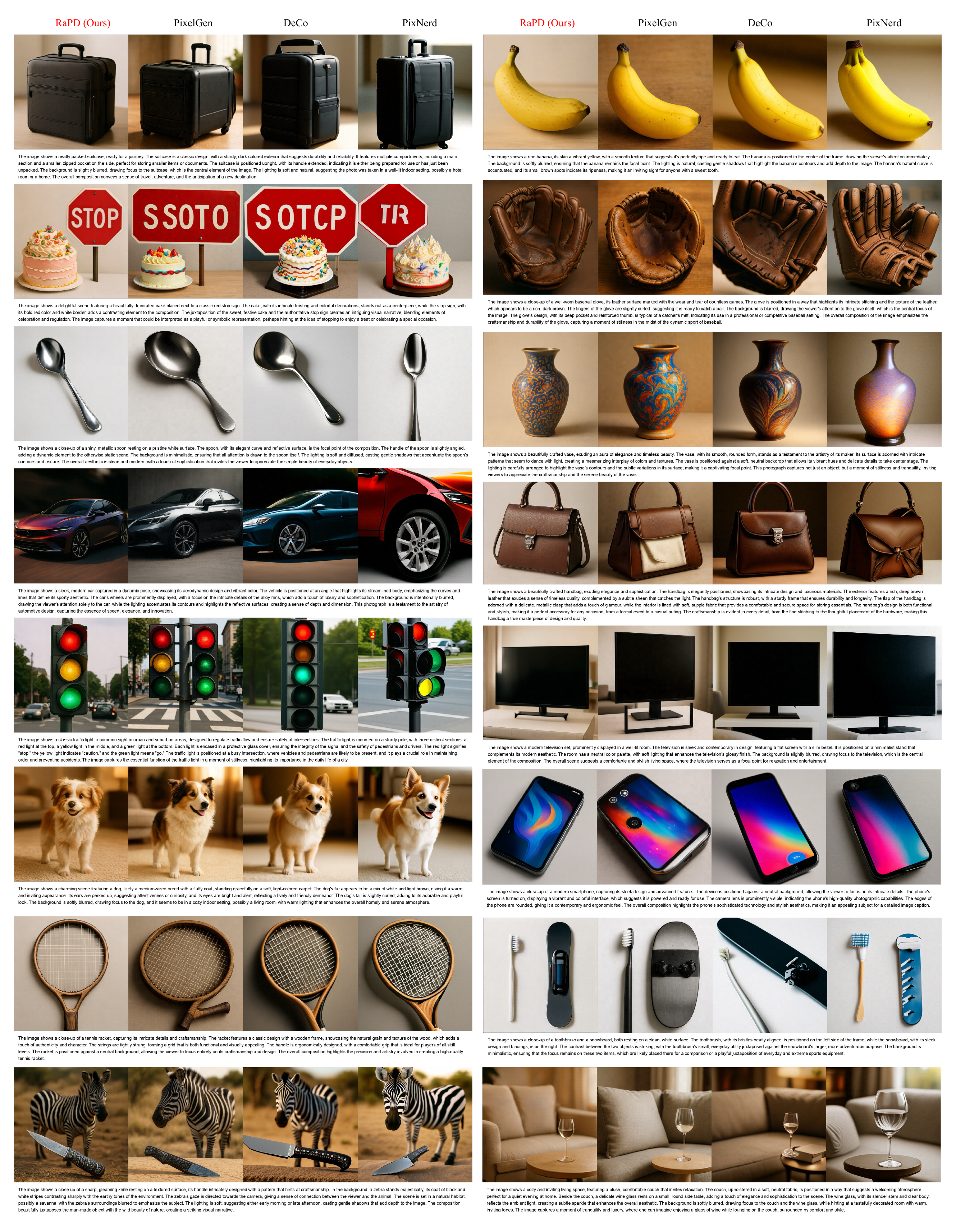}}
  \vspace{-0.6cm}
  \caption{Additional qualitative text-to-image comparisons between RaPD and baselines. Each row corresponds to a different prompt; each column shows the output of one method at the same target resolution. RaPD better preserves fine-grained detail and prompt fidelity, especially at higher resolutions.}
  \label{fig:comparison_appendix}
\end{figure}

To complement the qualitative results presented in the main paper, we provide additional side-by-side comparisons between RaPD and representative baselines on a broader set of prompts and target resolutions in Fig.~\ref{fig:comparison_appendix}. Across diverse content categories (objects, scenes, portraits, and text-rich compositions), RaPD consistently produces sharper textures, more coherent global structure, and stronger prompt alignment than competing methods. The advantage is particularly evident at higher target resolutions, where baseline approaches either degrade in fine-grained detail or fail to scale beyond their training grid, whereas RaPD maintains visual fidelity by simply varying the CQAR query coordinate grid over the same denoised latent.

\clearpage





\clearpage

\section*{NeurIPS Paper Checklist}

The checklist is designed to encourage best practices for responsible machine learning research, addressing issues of reproducibility, transparency, research ethics, and societal impact. Do not remove the checklist: {\bf The papers not including the checklist will be desk rejected.} The checklist should follow the references and follow the (optional) supplemental material.  The checklist does NOT count towards the page
limit. 

Please read the checklist guidelines carefully for information on how to answer these questions. For each question in the checklist:
\begin{itemize}
    \item You should answer \answerYes{}, \answerNo{}, or \answerNA{}.
    \item \answerNA{} means either that the question is Not Applicable for that particular paper or the relevant information is Not Available.
    \item Please provide a short (1--2 sentence) justification right after your answer (even for \answerNA). 
\end{itemize}

{\bf The checklist answers are an integral part of your paper submission.} They are visible to the reviewers, area chairs, senior area chairs, and ethics reviewers. You will also be asked to include it (after eventual revisions) with the final version of your paper, and its final version will be published with the paper.

The reviewers of your paper will be asked to use the checklist as one of the factors in their evaluation. While \answerYes{} is generally preferable to \answerNo{}, it is perfectly acceptable to answer \answerNo{} provided a proper justification is given (e.g., error bars are not reported because it would be too computationally expensive'' or ``we were unable to find the license for the dataset we used''). In general, answering \answerNo{} or \answerNA{} is not grounds for rejection. While the questions are phrased in a binary way, we acknowledge that the true answer is often more nuanced, so please just use your best judgment and write a justification to elaborate. All supporting evidence can appear either in the main paper or the supplemental material, provided in appendix. If you answer \answerYes{} to a question, in the justification please point to the section(s) where related material for the question can be found.

IMPORTANT, please:
\begin{itemize}
    \item {\bf Delete this instruction block, but keep the section heading ``NeurIPS Paper Checklist"},
    \item  {\bf Keep the checklist subsection headings, questions/answers and guidelines below.}
    \item {\bf Do not modify the questions and only use the provided macros for your answers}.
\end{itemize}


\begin{enumerate}

\item {\bf Claims}
    \item[] Question: Do the main claims made in the abstract and introduction accurately reflect the paper's contributions and scope?
    \item[] Answer: \answerYes{} 
    \item[] Justification: The abstract and introduction clearly summarize the main contributions, they are consistent with the methods and results represented in the paper.

\item {\bf Limitations}
    \item[] Question: Does the paper discuss the limitations of the work performed by the authors?
    \item[] Answer: \answerYes{} 
    \item[] Justification: see Sec.6

\item {\bf Theory assumptions and proofs}
    \item[] Question: For each theoretical result, does the paper provide the full set of assumptions and a complete (and correct) proof?
    \item[] Answer: \answerNA{} 
    \item[] Justification: The paper does not include any theoretical results, assumptions, or proofs.

    \item {\bf Experimental result reproducibility}
    \item[] Question: Does the paper fully disclose all the information needed to reproduce the main experimental results of the paper to the extent that it affects the main claims and/or conclusions of the paper (regardless of whether the code and data are provided or not)?
    \item[] Answer: \answerYes{} 
    \item[] Justification: The full method (encoder, CQAR, SRG, JiT denoiser) is described in Sec.~\ref{sec:method}; complete architecture and hyperparameter specifications are provided in Appendix Sec.~\ref{app:cqar}, \ref{app:denoiser}, and \ref{app:impl}, and training/data protocols are listed in Appendix Sec.~\ref{app:impl}.

\item {\bf Open access to data and code}
    \item[] Question: Does the paper provide open access to the data and code, with sufficient instructions to faithfully reproduce the main experimental results, as described in supplemental material?
    \item[] Answer: \answerYes{} 
    \item[] Justification: The training data (BLIP3o) is publicly available, and all evaluation benchmarks (GenEval, DPG-Bench) are public. The code and pretrained checkpoints will be released upon acceptance to facilitate reproducibility; we do not release them at submission time to preserve anonymity.

\item {\bf Experimental setting/details}
    \item[] Question: Does the paper specify all the training and test details (e.g., data splits, hyperparameters, how they were chosen, type of optimizer) necessary to understand the results?
    \item[] Answer: \answerYes{} 
    \item[] Justification: The experimental protocol is described in Sec.~\ref{sec:experiments}; full hyperparameters (learning rate, batch size, optimizer, EMA decay, training steps, precision, gradient clipping, sampling settings, CFG scale, and timestep shift) are summarized in Tab.~\ref{tab:model_spec} of Appendix Sec.~\ref{app:impl}.

\item {\bf Experiment statistical significance}
    \item[] Question: Does the paper report error bars suitably and correctly defined or other appropriate information about the statistical significance of the experiments?
    \item[] Answer: \answerYes{} 
    \item[] Justification: Following standard practice in large-scale text-to-image generation, our quantitative results (GenEval, DPG-Bench, rFID) are reported on the full evaluation sets shown in Tab.~\ref{tab:geneval} and Tab.~\ref{tab:dpg}

\item {\bf Experiments compute resources}
    \item[] Question: For each experiment, does the paper provide sufficient information on the computer resources (type of compute workers, memory, time of execution) needed to reproduce the experiments?
    \item[] Answer: \answerYes{} 
    \item[] Justification: Training/inference compute (training steps, batch size, precision, sampler steps) and resolution-vs-cost scaling are reported in Tab.~\ref{tab:model_spec} of Appendix Sec.~\ref{app:impl} and Tab.~\ref{tab:compute_scaling} of Sec.~\ref{sec:experiments}. The diffusion cost is fixed regardless of the output size, and only the CQAR decoding cost scales with the output token count.
    
\item {\bf Code of ethics}
    \item[] Question: Does the research conducted in the paper conform, in every respect, with the NeurIPS Code of Ethics \url{https://neurips.cc/public/EthicsGuidelines}?
    \item[] Answer: \answerYes{} 
    \item[] Justification: This work is foundational research on resolution-agnostic image generation; it does not involve human subjects, sensitive personal data, or deployments that violate the NeurIPS Code of Ethics. All third-party assets are used under their respective licenses.

\item {\bf Broader impacts}
    \item[] Question: Does the paper discuss both potential positive societal impacts and negative societal impacts of the work performed?
    \item[] Answer: \answerYes{} 
    \item[] Justification: Positive impacts include broader access to high-quality, resolution-flexible image synthesis with reduced compute, lowering the barrier for creative and educational applications. Potential negative impacts such as misuse for disinformation or non-consensual imagery are inherent to text-to-image generation in general; we discuss these in Sec.~\ref{sec:conclusion}.
    
\item {\bf Safeguards}
    \item[] Question: Does the paper describe safeguards that have been put in place for responsible release of data or models that have a high risk for misuse (e.g., pre-trained language models, image generators, or scraped datasets)?
    \item[] Answer: \answerYes{} 
    \item[] Justification: Upon release, we plan to distribute pretrained model weights with a usage license that prohibits malicious uses.

\item {\bf Licenses for existing assets}
    \item[] Question: Are the creators or original owners of assets (e.g., code, data, models), used in the paper, properly credited and are the license and terms of use explicitly mentioned and properly respected?
    \item[] Answer: \answerYes{} 
    \item[] Justification: All third-party assets used in this work are properly cited. We use each asset within its respective license.

\item {\bf New assets}
    \item[] Question: Are new assets introduced in the paper well documented and is the documentation provided alongside the assets?
    \item[] Answer: \answerYes{} 
    \item[] Justification: The newly introduced asset is the RaPD model. Its full architecture and training procedure are documented in Sec.~\ref{sec:method} and Appendix Sec.~\ref{app:cqar}, \ref{app:denoiser}, and \ref{app:impl}; on release, the model card and documentation will accompany the released checkpoints.

\item {\bf Crowdsourcing and research with human subjects}
    \item[] Question: For crowdsourcing experiments and research with human subjects, does the paper include the full text of instructions given to participants and screenshots, if applicable, as well as details about compensation (if any)? 
    \item[] Answer: \answerNA{} 
    \item[] Justification: This work does not involve crowdsourcing or research with human subject.

\item {\bf Institutional review board (IRB) approvals or equivalent for research with human subjects}
    \item[] Question: Does the paper describe potential risks incurred by study participants, whether such risks were disclosed to the subjects, and whether Institutional Review Board (IRB) approvals (or an equivalent approval/review based on the requirements of your country or institution) were obtained?
    \item[] Answer: \answerNA{} 
    \item[] Justification: This work does not involve research with human subjects, so IRB approval is not applicable.

\item {\bf Declaration of LLM usage}
    \item[] Question: Does the paper describe the usage of LLMs if it is an important, original, or non-standard component of the core methods in this research? Note that if the LLM is used only for writing, editing, or formatting purposes and does \emph{not} impact the core methodology, scientific rigor, or originality of the research, declaration is not required.
    \item[] Answer: \answerNA{} 
    \item[] Justification: We use Qwen3-1.7B as a frozen pre-trained text encoder, which is a standard component for text-conditioned image generation and not a novel methodological contribution. No LLM is used as a non-standard or original component of the proposed method.

\end{enumerate}

\end{document}